\documentclass[11pt]{article}

\PassOptionsToPackage{hyphens}{url}
\usepackage[final]{acl}

\usepackage{times}
\usepackage{latexsym}

\usepackage[T1]{fontenc}

\usepackage[utf8]{inputenc}

\usepackage{microtype}
\usepackage{inconsolata}

\usepackage{graphicx}

\usepackage{amsmath}
\usepackage{amssymb}
\usepackage{subcaption}
\usepackage{multirow}

\usepackage{graphicx}
\usepackage{makecell}
\usepackage{xcolor}
\usepackage[hyphens]{url}

\usepackage{comment}
\usepackage{booktabs}
\title{Where Knowledge Collides: A Mechanistic Study of Intra-Memory Knowledge Conflict in Language Models}

\newcommand*\samethanks[1][\value{footnote}]{\footnotemark[#1]}

\author{
  Minh Vu Pham \thanks{Equal Contribution}
  \quad
   Hsuvas Borkakoty \samethanks
  \quad
  Yufang Hou
  \\[4pt]
    IT:U Interdisciplinary Transformation University Austria
    \\
  \texttt{\{minhvu.pham, hsuvas.borkakoty, yufang.hou\}@it-u.at}
}

\begin{document}
\maketitle
\begin{abstract}

In language models (LMs), intra-memory knowledge conflict arises when inconsistent information about the same subject is encoded within the model's parametric knowledge \cite{xu-etal-2024-knowledge-conflicts}. Prior work has primarily focused on resolving conflicts between a model's internal knowledge and external sources, which is known as context-memory knowledge conflict, through approaches such as fine-tuning or knowledge editing, while the understanding of conflicts that arise internally remains largely unexplored. In this work, we design a framework to identify where internal conflicting knowledge is encoded within LMs. We test our framework on four LMs using both synthetic and real-world entities for knowledge conflicts. We find that internal conflicts often arise and are resolved in the final layers across all models, but that interventions are markedly less effective on real-world entity knowledge conflicts. Targeted attention-head interventions outperform layer-wise ones, and a filtering analysis shows that heads specialized for a single competing fact are far more common for synthetic entities, helping explain this gap. Finally, we find no evidence of universal attention heads dedicated to resolving knowledge conflicts. Instead, our results suggest that distinct circuits may separately encode competing pieces of knowledge, giving rise to conflict. Our results offer a first mechanistic account of intra-memory conflict resolution and highlight a substantial gap between synthetic and real-world settings.\footnote{Codes and datasets can be found at \url{https://github.com/ITU-NLP/where-knowledge-collides}}

\end{abstract}

\section{Introduction}
\label{sec:intro}



\begin{figure*}[t]
    \centering
    \begin{subfigure}[b]{0.36\textwidth}
        \centering
        \raisebox{3mm}{\includegraphics[width=\linewidth]{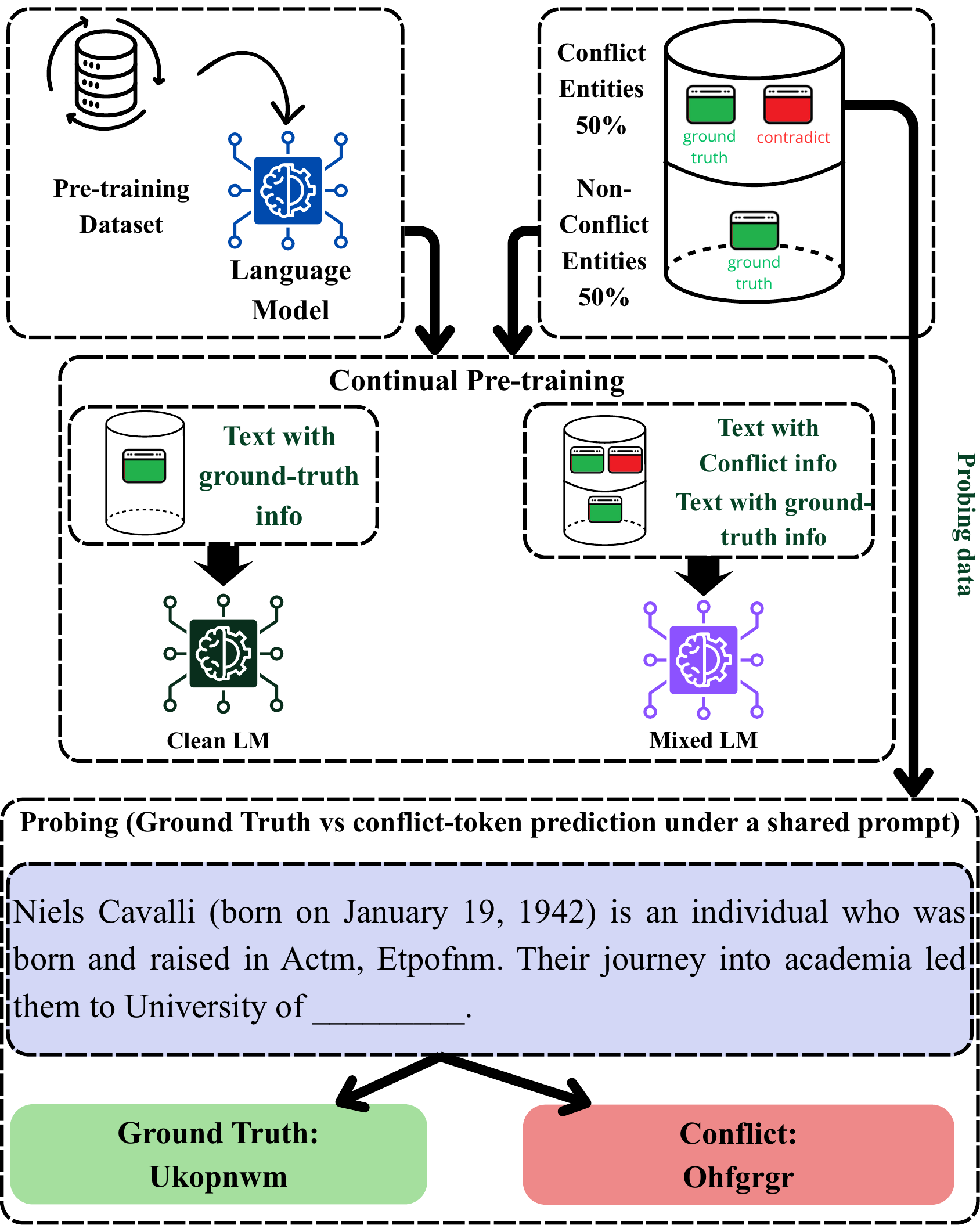}}
        \caption{}
        \label{fig:front-page}
    \end{subfigure}%
    \hfill
    \begin{subfigure}[b]{0.61\textwidth}
        \centering
        \includegraphics[width=\linewidth]{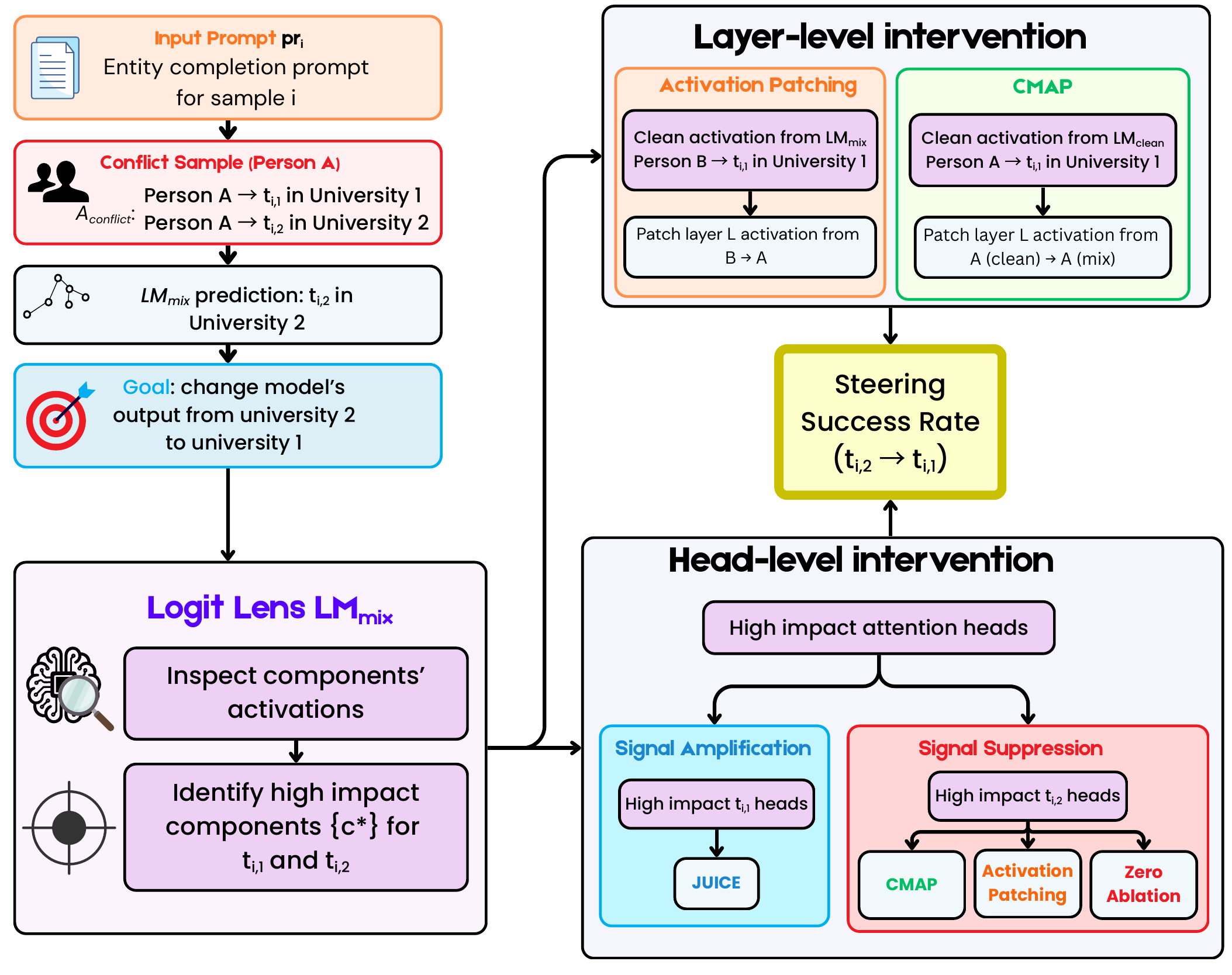}
        \caption{}
        \label{fig:pipeline_detailed}
    \end{subfigure}
    \caption{Overview of our proposed framework. (a) An overview of SynWikiBio with an example. The entity names are randomly generated to prevent pretrained data contamination. (b) The three-stage pipeline, outlining the experimental procedure used to localize and analyze the conflicting model components.}
    \label{fig:overview}
\end{figure*}

As information evolves rapidly, large language models (LLMs) face the challenge of keeping their parametric knowledge accurate to remain reliable and reduce hallucinations \cite{DBLP:journals/corr/abs-2402-01763}. Continual learning \cite{Ke2023ContinualPO} and retrieval-augmented generation \cite{DBLP:conf/nips/LewisPPPKGKLYR020,DBLP:conf/icml/BorgeaudMHCRM0L22} aim to keep model knowledge up to date, yet models can still encode contradictory information within their parameters, a phenomenon known as \emph{intra-memory knowledge conflict} \cite{xu-etal-2024-knowledge-conflicts}. Such conflicts compromise reliability by creating unstable internal representations, where different prompts or contexts elicit inconsistent factual associations and conflicting generations. They arise from biases and noise in training data \cite{DBLP:conf/fat/BenderGMS21, lazaridou2021mind}, limitations of knowledge editing \cite{li2023unveiling}, and sampling randomness in decoding \cite{DBLP:journals/tois/HuangYMZFWCPFQL25}, raising broader reliability concerns as LLM-backed systems become more integrated into daily use \cite{bengio2024managing,chatterji2025people,DBLP:journals/corr/abs-2506-11789}.

Existing mitigation methods focus on improving factual accuracy, including targeted fine-tuning \cite{Neeman2022DisentQADPA}, context-faithful prompting \cite{zhou-etal-2023-context}, and pruning internal components such as conflicting attention heads \cite{jin-etal-2024-cutting}. However, no prior work has addressed a fundamental question: \emph{where does knowledge conflict arise within LMs and which components are responsible for resolving it?} 

In this work, we propose a framework grounded in mechanistic interpretability (MI) to localize where internal conflicting knowledge is encoded. As illustrated in Figure~\ref{fig:front-page}, we construct \texttt{SynWikiBio}, a curated set of Wikipedia-style biographies for synthetic entities that deliberately contain conflicting facts (e.g., a person's graduating university), and use it to fine-tune a language model (e.g., GPT2-XL). We then probe the fine-tuned model with prompts designed to elicit competing knowledge representations (Section~\ref{sec:problemsetup}), apply the \emph{logit lens} to localize the components, such as attention layers or attention heads, that encode the conflict (Appendix~\ref{app:logit_len}), and assess their causal contribution via \emph{Activation Patching} (Section~\ref{subsec4:activation_patching}) and \emph{cross-model Activation Patching (CMAP)} \cite{DBLP:conf/iclr/PrakashSHBB24}, which enables cleaner causal attribution under conflicting knowledge (Section~\ref{subsec4:cross_model_patching}). Additionally, we use \texttt{DynamicQA} \cite{marjanovic-etal-2024-dynamicqa}, a knowledge conflict dataset containing real-world entities and events, to verify the generalizability of our findings.

This framework lets us investigate four research questions: (\textbf{Q1}) which components most affect LMs' decisions, and to what extent do intervention methods localize them? (\textbf{Q2}) how reliably can targeted patching and other interventions alter LMs' decisions in conflict settings? (\textbf{Q3}) how effective are patching activations across models sharing the same architecture but trained on different datasets? (\textbf{Q4}) how generalizable are our findings?

Across all experiments results of four LMs and two datasets (Section \ref{sec:results}), we summarize our findings as follows. 
First, conflict arises within the final few layers in every model. Comparing models with and without conflicting knowledge of a given subject reveals that both gradually converge toward a confident prediction until the final few layers, where the conflicting model grows uncertain while the non-conflicting model continues to converge toward its answer.
Second, targeted head-level interventions outperform layer-wise ones, though the effectiveness varies between different intervention methods and the number of intervened heads (the detailed results are shown in Sections~\ref{sec:steering_model_generation} and \ref{sec:target_intervention}).
Third, CMAP achieves results comparable to Activation Patching, suggesting it is a promising alternative for causal intervention in internal knowledge conflict settings, with a substantially simpler setup in real-world knowledge conflict settings.
Finally, the core findings such as conflict localization or better results of targeted heads over layer-wise intervention hold consistently across all four tested models spanning different scales and architectures, for both real and synthetic entity knowledge conflicts. However, the robustness of different intervention methods remains model-dependent.

In summary, our work provides initial evidence about where intra-memory knowledge conflicts emerge within LMs and how targeted interventions affect their resolution. Our findings also reveal important differences between synthetic and real-world entity conflicts, underscoring the need for further research on reliable methods to diagnose and mitigate internal knowledge conflicts.


\section{Related Work}

\paragraph{Intra-Memory Knowledge Conflict} refers to cases where a model generates different responses to similar inputs, primarily due to conflicting information in its training data \cite{xu-etal-2024-knowledge-conflicts}. While often considered similar to context-memory conflict, the two differ in certain characteristics: context-memory conflict allows a degree of control through additional context, and prior work, notably \citet{olsson2022context}, shows that In-Context Learning (ICL) exhibits unique behaviors (e.g., induction heads) absent in non-context settings. While numerous studies have addressed context-memory conflicts \cite[\textit{among others}]{Neeman2022DisentQADPA, jin-etal-2024-cutting, Ortu2024CompetitionOMA, DBLP:conf/icml/Li0T25}, very few have examined intra-memory conflicts, with only two notable studies to our knowledge: \citet{li-etal-2024-formality} and \citet{marjanovic-etal-2024-dynamicqa}. 
Specifically, \citet{marjanovic-etal-2024-dynamicqa} provides a real-world conflict dataset, \texttt{DynamicQA}, and shows that LMs exhibit greater uncertainty on disputable facts than on canonical ones. \citet{li-etal-2024-formality} shows that LMs develop human-like biases toward formal, well-written text with fewer spelling errors. However, an in-depth understanding of how LMs process knowledge conflicts internally has yet to be explored. We address this gap by developing a mechanistic interpretability-based framework to localize conflicting knowledge encoded within language models.

\paragraph{Mechanistic Interpretability (MI)} is the field in machine learning that focuses on understanding the internal mechanisms of neural networks \cite{olah2020zoom,bereska2024mechanistic}. Prior work has applied  \emph{logit lens} \cite{Ortu2025WhenSO} and \emph{Activation Patching} \cite{DBLP:conf/iclr/ZhangN24} for a range of analyses and interventions, including explanation of how LMs handle context-memory conflict  \cite{Ortu2024CompetitionOMA} and analysis of knowledge selection behavior in LMs \cite{Zhao2024SteeringKSA}. In this work, we apply both techniques to localize and causally intervene on conflicting knowledge encoded in language models that originates from the training data.

\section{Experiment Setup}
\label{sec:expsetup}

\subsection{Problem Definition}
\label{sec:problemsetup}
We formalize the problem as follows. Let $\mathcal{B}$ denote a dataset comprising personal biographies. A biography $b_i \in \mathcal{B}$ is said to contain knowledge conflict if there exists another biography $\bar{b}_i \in \mathcal{B}$ such that $b_i$ and $\bar{b}_i$ refer to the same individual $p_i$ and share identical personal attributes except for one conflicting element. On the other hand, a conflict-free instance $b_i \in \mathcal{B}$ means that $b_i$ is the only biography in $\mathcal{B}$ that describes person $p_i$. 

First, we construct dataset $\mathcal{B}_{mix}$, which comprises both conflict-containing and conflict-free personal biographies.
Let $\ell$ be a pretrained language model, we continue pretraining $\ell$ on $\mathcal{B}_{mix}$, yielding model $\ell_{mix}$. It is worth noting that this process applies to our synthetic dataset, \texttt{SynWikiBio}, but can also be extended to \texttt{DynamicQA} by treating each factual context in \texttt{DynamicQA} as a biography. Further dataset details are presented in Section~\ref{sec:datasets}.

Following established mechanistic interpretability studies \cite{elhage2021mathematical, DBLP:conf/nips/0001LV23, Ortu2024CompetitionOMA}, we formulate an entity completion task on model $\ell_{mix}$ that explicitly elicits conflicting parametric knowledge as illustrated in Figure \ref{fig:front-page}. Specifically, for each person $p_i$ whose biography contains conflicting information, we apply $p_i$'s name and related information to a randomly selected predefined template (full list of templates is included in appendix \ref{sec:templates}) to create prompt $pr_i$ that induces the conflict information from $\ell_{mix}$'s parametric knowledge (e.g., ``Niels Cavalli studies at University of \_\_\_''). 
The process of recalling both knowledge internally by $\ell_{mix}$ can be traced by investigating the logits of these knowledge's first tokens, which by construction  uniquely identify each entity (Appendix~\ref{app:dataset}). We verify consistency with the full generated entity.
For future reference, we use $t_{i,1} \in b_i$, $t_{i,2} \in \bar{b}_i$ as a reference to the first tokens of the conflicting attribute, for any individual $p_i$ in $\mathcal{B}_{mix}$ whose biography contains a contradictory factual claim per definition.


\subsection{Dataset}
\label{sec:datasets}

For this study, we create a synthetic dataset of personal biographies ($\mathcal{B}$) in the style of Wikipedia short biographies, consisting of fictional individuals each with a unique set of personal attributes: \textit{birth date}, \textit{birth place}, \textit{university}, \textit{major}, \textit{company}, and \textit{work place} \cite{DBLP:conf/icml/Allen-ZhuL24, li-etal-2024-formality, zucchet2025how}. We follow \citet{li-etal-2024-formality}'s generation process with GPT-4o \cite{hurst2024gpt}, applying a randomly selected template to each biography to promote generalization. Entity names for the important attributes \textit{birth place}, \textit{company}, and \textit{university} are replaced with randomly generated words, while less relevant attributes (\textit{major}, \textit{work place}) remain unchanged; the conflicting attribute is always in either the \textit{university} or \textit{company} category.
From $\mathcal{B}$, we construct $\mathcal{B}_{mix}$ (\texttt{SynWikiBio}), containing $n_1 = 1000$ people with knowledge-conflicting biographies and $n_2 = 1000$ people with conflict-free biographies. We further derive $\mathcal{B}_{clean}$ (\texttt{SynWikiBio\_clean}) from $\mathcal{B}_{mix}$ by removing $\bar{b}_i$ from each conflicting pair $(b_i,\ \bar{b}_i)$, yielding a conflict-free dataset. Continued pretraining of $\ell$ on $\mathcal{B}_{clean}$ gives model $\ell_{clean}$. Dataset generation details and statistics are in Appendix \ref{app:dataset}. 

To validate the generalizability of our findings to conflicts in real-world entities, we use \texttt{DynamicQA} \cite{marjanovic-etal-2024-dynamicqa}, a real-world knowledge-conflict dataset containing \textbf{Static} facts (unchanging over time), \textbf{Temporal} facts (changing over time), and \textbf{Disputable} facts (varying by viewpoint) across a wide range of subjects. Each fact is provided with a context, which we treat as an equivalent to a biography in \texttt{SynWikiBio}. We construct conflicting contexts by replacing the factual knowledge with the counterfactual knowledge \texttt{DynamicQA} provides. We then randomly select 2000 contexts, 1000 of which are conflicting, to continue pretraining $\ell_{mix}$, and remove the conflicting contexts to obtain $\ell_{clean}$. Unlike \texttt{SynWikiBio}, where we evaluate the first generated token, for \texttt{DynamicQA} we compare the fully generated entities.

\subsection{Models}
\label{sec:model}

We evaluate our framework on four language models, GPT-2 XL 1.5B \cite{radford2019language}, Qwen3-4B \cite{qwen3}, OPT-2.7B \cite{zhang2022opt} and Llama-3.2-1B \cite{llama32}.
These four models cover both widely studied architecture that serves as a standard baseline in mechanistic interpretability research, and models of different sizes to test generalizability across scales. We use Transformers library \cite{wolf-etal-2020-transformers} for training the models, and TransformerLens \cite{nanda2022transformerlens} for causal tracing tasks. The architectural differences and implementation details are reported in Appendix \ref{app:implementation}.

\section{Causal Probing Intra-Memory Knowledge Conflict}
\label{sec4:logit_activation_theory}

As shown in Figure~\ref{fig:pipeline_detailed}, we develop a framework based on $\ell_{\text{mix}}$ and $\ell_{\text{clean}}$ to probe the model's parametric knowledge. First, we perform logit inspection \cite{nostalgebrist2020logitlens} on the residual stream of $\ell_{\text{mix}}$ to localize components responsible for encoding and generating conflicting information from \texttt{SynWikiBio} or \texttt{DynamicQA}~(Appendix~\ref{app:logit_len},~\ref{app:head_contrib_compute}).  
We then apply Activation Patching \cite{DBLP:conf/iclr/ZhangN24} within $\ell_{\text{mix}}$ to test whether these components causally influence generation with respect to the conflicting facts (Section~\ref{subsec4:activation_patching}). Finally, we experiment with cross-model Activation Patching, an advanced causal tracing method that leverages both $\ell_{\text{mix}}$ and $\ell_{\text{clean}}$ to analyze intra-memory knowledge conflicts (Section~\ref{subsec4:cross_model_patching}).

\subsection{Preliminaries}
\label{subsec4:prelim}

\paragraph{Logit lens}
is a technique for inspecting a model’s intermediate representations by decoding the residual stream at each layer, thereby revealing how model's prediction evolves across different components \cite{nostalgebrist2020logitlens}. 

\paragraph{Activation patching} identifies task-relevant components by replacing a component's activation under a target prompt with its activation from an alternative (source) prompt and measuring the resulting change in model behavior \cite{DBLP:conf/nips/MengBAB22, DBLP:conf/iclr/ZhangN24}. Given source prompt $pr_j$ and target prompt $pr_i$, forward passes $\ell_{mix}(pr_j)$ and $\ell_{mix}(pr_i)$ yield activation sets $A_j = \{a^{l,\, c}_j \mid l \in L,\, c \in \{\text{attn},\, \text{mlp}\}\}$ and $A_i$, respectively, where $L$ is the number of layers in $\ell_{mix}$. To measure the causal effect of component $c$ at layer $l$, we run $\ell_{mix}(pr_i)$ but replace its activation $a^{l,\ c}_i$ with the corresponding source activation $a^{l,\ c}_j$, recomputing all downstream activations, denoted $\ell_{mix}(pr_i \mid a^{l,\ c}_i \leftarrow a^{l,\ c}_j)$. Impact is measured by the output difference between this patched run and the original $\ell_{mix}(pr_i)$.

\paragraph{Zero ablation} zeroes out component's activation to measure its contribution to the model's output.

\paragraph{JUICE} \cite{DBLP:conf/icml/Li0T25} is an intervention method that runs the model twice. The first pass caches activations of the attention heads contributing most to a target token, the second pass patches these activations back, scaled by a factor $\alpha$, to steer generation toward the desired source.

\subsection{Causal Tracing via Activation Patching}
\label{subsec4:activation_patching}
After localizing components strongly associated with conflicting information for a given attribute type (e.g., \emph{university}) in $\ell_{\text{mix}}$, we apply Activation Patching to causally assess each component $c$'s contribution to the model's output behavior.
For each sample $b_i$ (biography in \texttt{SynWikiBio}, context in \texttt{DynamicQA}) containing knowledge conflict, we quantify the causal effect of component $c$ using target prompt $pr_i$, which induces competition between two next-token continuations $t_{i,1}$ and $t_{i,2}$, corresponding to $b_i$ and $\bar{b}_i$ respectively (Section~\ref{sec:problemsetup}). We select two samples $b_j$ from the conflict-free set, yielding two source prompts $pr_j$, chosen such that the forward passes $\ell_{mix}(pr_j)$ produce $t_{i,1}$ and $t_{i,2}$ respectively.
Let $P_t(\cdot)$ and $P'_t(\cdot)$ denote the model's probability for token $t$ before and after Activation Patching. The causal effect of component $c$ at layer $l$ on generating $t = \{t_{i,1},\ t_{i,2}\}$ under prompt $pr_i$ is defined as:
\[
\Delta^{l,\ c}_{t}
\;:=\;
P'_{t}
\;-\;
P_{t}
\]
The forward pass $\ell_{mix}(pr_i \mid a^{l,\ c}_i \leftarrow a^{l,\ c}_j)$ is patched at the last input token position, immediately preceding generation of $t$, isolating $c$'s causal influence on resolving factual conflicts by measuring how substituting a clean, fact-consistent activation shifts the model's preference between competing continuations.

\subsection{Cross Model Activation Patching}
\label{subsec4:cross_model_patching}

Standard Activation Patching sources activations from prompts processed by the same model, implicitly assuming the availability of a similar yet non-confounded source prompt. This assumption fails under intra-memory knowledge conflict: a source prompt $pr_j$ whose biography $b_j$ is absent from training may fail to elicit faithful recall, while one too similar to the target biography $b_i$ risks being conflated with it, causing the model to encode overlapping or competing internal representations. In either case, the model either fails to retrieve the intended fact or cannot cleanly separate source and target memories, yielding unstable activations.
Activation Patching mitigates the problem by selecting source prompts $pr_j$ that share the target attribute value but correspond to a different individual $p_j$. While this reduces prompt-target entanglement, it introduces noise from inter-person variation, weakening causal attribution under factual conflict.

To resolve this, we adopt \emph{cross-model Activation Patching (CMAP)} \cite{DBLP:conf/iclr/PrakashSHBB24}, tailored here to intra-memory conflict. Rather than sourcing activations from alternative prompts within the conflicted model $\ell_{\text{mix}}$, CMAP transfers activations from a \emph{clean} reference model $\ell_{\text{clean}}$, obtained by continuing pretraining the base model $\ell$ on \texttt{SynWikiBio\_clean}, which contains one conflict-free biography per individual and no contradictory claims. 
During patching, activations from $\ell_{\text{clean}}$ are injected into $\ell_{\text{mix}}$ at corresponding components and layers while processing the conflicting prompt $pr_i$. The two models share the same base initialization and the same set of individuals $\{p_i\}$, differing only in the presence of contradictory biography $\bar{b}_i$. This alignment eliminates confounds from prompt mismatch and inter-person variation, ensuring that patched activations reflect uncontaminated parametric representations of $b_i$ and enabling more reliable causal attribution of the components involved in resolving factual conflicts within $\ell_{\text{mix}}$.

\section{Results and Findings}
\label{sec:results}

Section~\ref{sec:layer_wise_logit_lens} and Section~\ref{sec:model_uncertainty} report results from GPT2-XL on \texttt{SynWikiBio}, while the in-depth causal analyses from Section~\ref{sec:steering_model_generation} to \ref{sec:circuit} report results and comparisons across all tested LMs on both datasets.

\subsection{Layer-wise Logit Lens Probing Results}
\label{sec:layer_wise_logit_lens}

The results of the experiment investigating aggregate contributions across all $n_1$ individuals in $\ell_{mix}$ and across all prompts eliciting conflicting tokens $t_{i,1}$ or $t_{i,2}$ (Figure~\ref{fig:component_layerwise_analysis}, Appendix~\ref{app:logit_len}) show stronger aggregate contributions in later layers, particularly between layers 21 and 47. Consistent with prior studies~\cite{DBLP:conf/nips/MengBAB22, jin-etal-2024-cutting, DBLP:conf/icml/Li0T25}, this phenomenon is more pronounced in attention components than their corresponding MLP components across all layers. Accordingly, we focus our analysis on attention components at the later layers (Section \ref{sec:model_uncertainty} to \ref{sec:target_intervention}). 

\subsection{Layer-wise Confidence Analysis}
\label{sec:model_uncertainty}

\begin{figure}[t]
\includegraphics[width=\columnwidth]{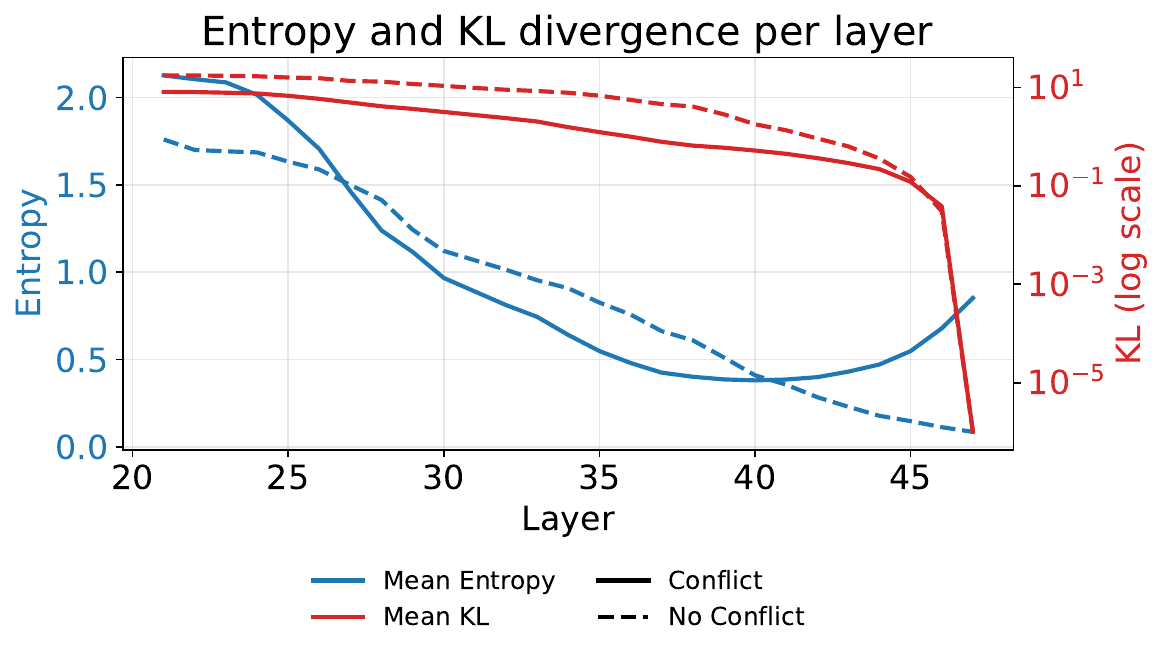}
\caption{GPT2-XL's confidence across layers 21-47, measured by entropy and KL divergence to the final layer distribution. Low entropy indicates high-confidence prediction at the current layer, while KL divergence quantifies how much the current predicted distribution differs from the model's final output.}
\label{fig:entropy_gpt2}
\end{figure}

To track how prediction confidence evolves across layers, we apply the logit lens to the residual stream at each layer of $\ell_{mix}$ and $\ell_{clean}$, computing next-token entropy and KL divergence (relative to the final-layer distribution) at each layer. Figure~\ref{fig:entropy_gpt2} shows the resulting averages. Entropy decreases synchronously for both models through the middle layers and diverges only at layer 40. From that point, $\ell_{clean}$ continues toward 0, while $\ell_{mix}$ rises toward 1, indicating that non-trivial probability mass shifts onto competing candidate tokens. This suggests the model commits to a prediction early and only encounters conflicting parametric knowledge in the final layers (40-47). KL divergence, by contrast, remains largely similar across layers for both models and proves less informative. This pattern is consistent across all four LMs tested (other results are shown in Appendix~\ref{sec:add_uncertainty}).

\subsection{Causal Intervention with Layer-wise Activation Patching and CMAP}
\label{sec:steering_model_generation}

In this experiment, we quantify the impact of attention layers through the rate at which patching flips the model's output (Steering Success Rate or SSR). A flip indicates that the intervened layer carries sufficient influence to affect how the model resolves conflicting information. We retain only samples for which $\ell_{\text{mix}}$ outputs $t_{i,2} \in \bar{b}_i$, since CMAP relies on $\ell_{\text{clean}}$ being fine-tuned only on $b_i \ni t_{i,1}$ and is therefore meaningful only when steering away from $t_{i,2}$ toward $t_{i,1}$. For both \texttt{SynWikiBio} (syn) and \texttt{DynamicQA} (dqa), we follow the next-token prediction task described in Section~\ref{sec:problemsetup}. For \texttt{DynamicQA}, we intervene using CMAP only, since Activation Patching requires a separate, non-conflicting sample that also produces $t_{i,1}$, which is difficult to find in a real-world dataset due to a lack of controlled conflicts/non-conflicts. This also highlights the value of \texttt{SynWikiBio}, which offers controlled conflicts for in-depth analysis, and CMAP, which requires a much easier setup than Activation Patching in complex settings such as knowledge conflict.

\begin{table}[h]
  \centering
  \footnotesize
  \setlength{\tabcolsep}{3pt}
  \resizebox{\columnwidth}{!}{%
  \begin{tabular}{@{}lccccc@{}}
    \toprule
     & \textbf{AP (Syn)} & \multicolumn{2}{c}{\textbf{CMAP (Syn)}} & \multicolumn{2}{c}{\textbf{CMAP (DQA)}} \\
    \cmidrule(lr){3-4} \cmidrule(lr){5-6}
     & & \textbf{top-syn} & \textbf{top-dqa} & \textbf{top-syn} & \textbf{top-dqa} \\
    \midrule
    GPT2-XL      & 0.235 (L43) & 0.214 (L44) & 0.208 (L46) & 0.094 (L44) & 0.141 (L46) \\
    Llama-1B     & 0.726 (L14) & 0.619 (L14) & - & - & 0.425 (L14) \\
    OPT-2.7B     & 0.649 (L28) & 0.665 (L28) & - & - & 0.303 (L28) \\
    Qwen3-4B     & 0.626 (L33) & 0.586 (L35) & 0.459 (L33) & 0.146 (L35) & 0.274 (L33) \\
    \bottomrule
  \end{tabular}%
  }
  \caption{Best steering success rate (SSR) for layer-wise Activation Patching, by patching type (with Layer number of each corresponding value shown within brackets). \textbf{top-syn} and \textbf{top-dqa} denote the highest impact layer for \texttt{SynWikiBio} and \texttt{DynamicQA}, respectively. Only flips from $t_2$ to $t_1$ are counted. Sample counts (AP/CMAP-syn versus CMAP-real) were: GPT2-XL 486/498, Llama-3.2-1B 493/468, OPT-2.7B 502/491, and Qwen3-4B 503/486.}
  \label{tab:layerwise-patching}
\end{table}

Table~\ref{tab:layerwise-patching} reports the top results, including CMAP performance at the \textbf{top-syn} and \textbf{top-dqa} layers for \texttt{SynWikiBio} and \texttt{DynamicQA}, respectively, for comparison.\footnote{Llama-3.2-1B and OPT-2.7B share the same best layer for both datasets, so no separate value is reported.} Across all four models, layer-wise CMAP shows a consistent drop in SSR when moving from synthetic to real-world conflicts, ranging from 7.3 points for GPT2-XL (0.214 to 0.141) and 19.4 points for Llama-3.2-1B (0.619 to 0.425) to 31.2 and 36.2 points for Qwen3-4B (0.586 to 0.274) and OPT-2.7B (0.665 to 0.303), respectively, indicating that real-world knowledge conflicts are inherently harder to overwrite via patching. For all tested LMs, the highest-impact layer falls within the last few layers for both datasets, reaffirming our assumption in Section~\ref{sec:model_uncertainty} that conflicts arise toward the end of the network.

Although entity distributions between $\ell_{clean}$ and $\ell_{mix}$ differ slightly due to the absence of contradictory claims $\bar{b}_i$, an additional experiment (Appendix~\ref{app:cmap_freq}) shows that patching activations from $\ell_{clean}$ and from an auxiliary model $\ell'_{clean}$, trained with the same entity distribution as $\ell_{mix}$, yields near-identical results.

\begin{table*}[t]
\centering
\footnotesize
\setlength{\tabcolsep}{4pt}
\renewcommand{\arraystretch}{0.92}
\resizebox{\textwidth}{!}{%
\begin{tabular}{ll cccc cccc}
\toprule
& & \multicolumn{4}{c}{\textbf{SynWikiBio}} & \multicolumn{4}{c}{\textbf{DynamicQA}} \\
\cmidrule(lr){3-6} \cmidrule(lr){7-10}
Model & \makecell{Intervention\\Type} & \makecell{Num\\Samples} & \makecell{k-heads\\SSR} & \makecell{Avg Heads\\Post-filter} & \makecell{Filtered\\SSR ($\Delta$)} & \makecell{Num\\Samples} & \makecell{k-heads\\SSR} & \makecell{Avg Heads\\Post-filter} & \makecell{Filtered\\SSR ($\Delta$)} \\
\midrule
\multirow{4}{*}{GPT2-XL}
 & Ablation       & 474 & 0.646 & 13.40 & 0.669 \textcolor{green!60!black}{(0.023)}  & 488 & 0.434 & 10.63 & 0.410 \textcolor{red}{(0.025)} \\
 & Juice          & 474 & \textbf{0.960} & 13.40 & \textbf{0.983} \textcolor{green!60!black}{(0.023)}  & 488 & \textbf{0.777} & 10.63 & \textbf{0.777} (0.000) \\
 & CMAP           & 486 & 0.681 & 13.40 & 0.665 \textcolor{red}{(0.017)}             & 498 & 0.478 & 10.63 & 0.396 \textcolor{red}{(0.082)} \\
 & Act.\ Patching & 486 & 0.772 & 13.40 & 0.718 \textcolor{red}{(0.054)}             & --  & --    & --    & -- \\
\midrule
\multirow{4}{*}{LLaMa-3.2-1B}
 & Ablation       & 493 & 0.432 & 7.51 & 0.840 \textcolor{green!60!black}{(0.408)}   & 452 & 0.462 & 4.99 & 0.409 \textcolor{red}{(0.053)} \\
 & Juice          & 493 & 0.915 & 7.51 & \textbf{0.988} \textcolor{green!60!black}{(0.073)}   & 452 & \textbf{0.573} & 4.99 & \textbf{0.732} \textcolor{green!60!black}{(0.159)} \\
 & CMAP           & 493 & 0.805 & 7.51 & 0.801 \textcolor{red}{(0.004)}              & 468 & 0.526 & 4.99 & 0.340 \textcolor{red}{(0.186)} \\
 & Act.\ Patching & 493 & \textbf{0.959} & 7.51 & 0.846 \textcolor{red}{(0.114)}              & --  & --    & --   & -- \\
\midrule
\multirow{4}{*}{OPT-2.7B}
 & Ablation       & 502 & 0.476 & 13.93 & 0.994 \textcolor{green!60!black}{(0.518)} & 481 & 0.516 & 7.09 & 0.511 \textcolor{red}{(0.004)} \\
 & Juice          & 502 & \textbf{0.994} & 13.93 & \textbf{0.998} \textcolor{green!60!black}{(0.004)} & 481 & \textbf{0.684} & 7.09 & \textbf{0.871} \textcolor{green!60!black}{(0.187)} \\
 & CMAP           & 502 & 0.978 & 13.93 & 0.994 \textcolor{green!60!black}{(0.016)} & 491 & 0.611 & 7.09 & 0.466 \textcolor{red}{(0.145)} \\
 & Act.\ Patching & 502 & 0.986 & 13.93 & 0.994 \textcolor{green!60!black}{(0.008)} & --  & --    & --   & -- \\
\midrule
\multirow{4}{*}{Qwen3-4B}
 & Ablation       & 503 & 0.425 & 7.34 & 0.817 \textcolor{green!60!black}{(0.392)} & 472 & 0.331 & 5.29 & 0.203 \textcolor{red}{(0.127)} \\
 & Juice          & 503 & 0.839 & 7.34 & \textbf{0.970} \textcolor{green!60!black}{(0.131)} & 472 & \textbf{0.483} & 5.29 & \textbf{0.555} \textcolor{green!60!black}{(0.072)} \\
 & CMAP           & 503 & 0.921 & 7.34 & 0.799 \textcolor{red}{(0.121)}            & 486 & 0.354 & 5.29 & 0.152 \textcolor{red}{(0.202)} \\
 & Act.\ Patching & 503 & \textbf{0.984} & 7.34 & 0.867 \textcolor{red}{(0.117)}            & --  & --    & --   & -- \\
\bottomrule
\end{tabular}%
}
\caption{Filtered vs.\ non-filtered intervention success rates. \textbf{k-heads SSR} reports the SSR of interventions targeting the full $k=30$ heads, while \textbf{Filtered SSR} columns report SSR of interventions targeting filtered heads (those contributing strongly to only $t_{i,1}$ or $t_{i,2}$). $\Delta$ denotes the difference between Filtered SSR and k-heads SSR: \textcolor{green!60!black}{green} = improvement, \textcolor{red}{red} = decline. Bold indicates the best value in each column within a model's block. No Activation Patching interventions on \texttt{DynamicQA}.}
\label{tab:filtered-vs-original}
\end{table*}

\subsection{Attention-Head Targeted Intervention}
\label{sec:target_intervention}


Layer-wise intervention can be coarse, as individual attention heads within a single attention component often exhibit distinct behaviors. To identify high-impact attention heads, we first compute each head's contribution (detailed in Appendix~\ref{app:head_contrib_compute}) and select those contributing most strongly to $t_{i,1}$ and $t_{i,2}$, respectively. In addition to Activation Patching and CMAP, we employ two further intervention methods: zero ablation and JUICE (Just Run Twice; \citealp{DBLP:conf/icml/Li0T25}).

For each conflicting biography $b_i$ in the set of $n_1$ samples, we first determine the model's output on $pr_i$. Similar to Section~\ref{sec:steering_model_generation}, we retain only samples for which $\ell_{\text{mix}}$ outputs $t_{i,2} \in \bar{b}_i$. As tested LMs contain 25-32 heads per attention layer, we select the top $k=30$ heads (for each $t_{i,1}$, $t_{i,2}$) for fair comparison. 

For each retained sample, the intervention applied depends on the method. For Activation Patching, CMAP, and zero ablation, we patch or zero out the heads contributing most to $t_{i,2}$ to suppress $t_{i,2}$ and promote $t_{i,1}$ (Activation Patching and CMAP). For JUICE, we instead amplify the heads contributing most to $t_{i,1}$ by a factor $\alpha = 5$.

Table~\ref{tab:filtered-vs-original} reports the results for this 
attention head intervention experiment in \textbf{k-heads SSR} columns. We can see that targeted heads intervention shows much better results in \texttt{SynWikiBio}, with JUICE achieving from 0.839 with Qwen3-4B up to 0.994 with OPT-2.7B. Targeted-heads Activation Patching and CMAP also show better results compared to layer-wise intervention. This also holds for real-world conflict performance, as these interventions also outperform layer-wise intervention. 


Additionally, from the $k=30$ selected heads for each $t_{i,1}$ and $t_{i,2}$, we filter out heads that contribute highly to both tokens, leaving only heads that support only $t_{i,1}$ or $t_{i,2}$. The average number of remaining heads per sample (reported in \textbf{Avg Heads Post-filter} columns) is consistently higher in the synthetic setting. This indicates that real-world conflicts are often more complicated due to uncontrolled occurrences of $t_{i,1}$ and $t_{i,2}$ in the pre-training data, which leads to fewer decision-making heads. Consequently, isolating token-specific heads is inherently harder for real-world entities. Intervention with filtered heads yields consistent improvements for JUICE across all LMs and both datasets. Zero ablation also gains substantially in the synthetic setting (+0.408 for Llama-3.2-1B, +0.518 for OPT-2.7B, +0.392 for Qwen3-4B). This increase suggests that targeting specialized heads, those supporting only $t_{i,1}$ or $t_{i,2}$, is more effective for signal-amplifying/suppressing interventions. On the other hand, Activation Patching and CMAP likely rely on a broader set of heads to faithfully reconstruct the token-specific signal, including some of the heads that the filtering process removes. Checking the distribution of the selected heads (Appendix~\ref{app:head_dist}) also confirms knowledge conflicts in the final layers.

\subsection{Impact of Attention Heads}
\label{sec:head_vs_layer}


To understand the contribution of individual attention heads relative to full attention layers, we restrict patching to the top 5 heads per layer on \texttt{SynWikiBio}. This restriction consistently degrades performance relative to patching the entire layer: the best AP/CMAP scores drop to 0.146/0.108 for GPT2-XL (L44), 0.584/0.449 for Llama-3.2-1B (L14), 0.562/0.494 for OPT-2.7B (L28), and 0.427/0.318 for Qwen3-4B (L33). In contrast, patching the top $k=30$ heads selected across layers, roughly the size of a single layer, substantially outperforms single-layer patching (Table~\ref{tab:filtered-vs-original}). Taken together, these results suggest that the conflict resolving decision is distributed across many heads, and patching a handful of top heads is insufficient to steer the model's output. However, patching a comparable number of well-selected heads drawn from multiple layers is more effective than any single layer.

\subsection{Why are GPT2-XL's results worse?}
\label{sec:gpt2}

Throughout the experiments, we notice that CMAP and Activation Patching often perform worse with GPT2-XL than with the other LMs. This raises the question: \textit{why are interventions less effective on GPT2-XL?} To investigate, we select heads ranging from the top 1\% to top 15\% (by contribution) of the total heads in each LM and intervene using CMAP. We test all four LMs on both real-world and synthetic conflict settings, adding randomly selected heads of the same size in the real-world setting as a baseline. 

\begin{figure*}[t]
    \centering
    \resizebox{\textwidth}{!}{%
    \includegraphics{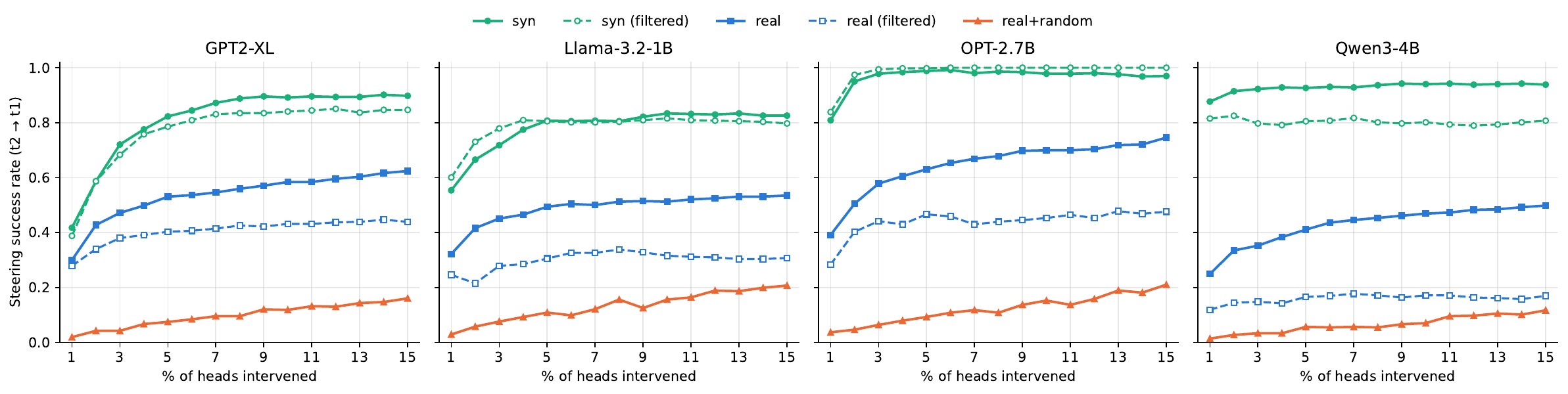}%
    }
   \caption{Steering success rate (SSR) of the percentage of attention heads intervened via CMAP (1\% to 15\%), for real-world (DynamicQA) and synthetic (SynWikiBio) entities knowledge conflicts across all four models. "real+random" denotes a baseline intervention with randomly selected heads with the same size, evaluated on \texttt{DynamicQA}. All models show a saturating trend, with GPT2-XL following the same pattern as the others once its larger total head count is accounted for. Number of samples (real / real+random; syn): GPT2-XL 524; 508, Llama-3.2-1B 488; 493, OPT-2.7B 519; 502, Qwen3-4B 514; 503.}
   \label{fig:gpt2_investigation}
\end{figure*}

The results from Figure~\ref{fig:gpt2_investigation} and additional analysis from Appendix \ref{app:gpt2} suggest two things: First, CMAP with randomly selected heads is consistently below other lines of CMAP with filtered and non-filtered on \texttt{DynamicQA} dataset. This shows that the selected heads indeed play a role in resolving internal knowledge conflicts.
Second, GPT2-XL is not an outlier. Its trend closely resembles that of Llama-3.2-1B, with SSR rising sharply from 1\% to 5\% of intervened heads before flattening out. GPT2-XL's weaker results so far stem from it having the largest total number of attention heads (i.e., 1200) among all tested LMs, meaning proportionally more heads must be intervened to match Llama-3.2-1B's performance. The result also suggests that, unlike Qwen3-4B or OPT-2.7B, which have more concentrated important heads (both have flattening performance after intervening more than top 2\% of heads), GPT2-XL distributes prediction-relevant computation across a much larger set of heads. Our earlier interventions, which used a small and fixed number of heads, were therefore insufficient to meaningfully affect GPT2-XL's predictions. These results also support the findings of Section~\ref{sec:head_vs_layer}, as we can see patching effectiveness increases with the number of patched heads for all tested models.

\subsection{Universal Circuit for Knowledge Conflict}
\label{sec:circuit}

\begin{table}[t]
  \centering
  \footnotesize
  \setlength{\tabcolsep}{4pt}
  \begin{tabular}{@{}lccc@{}}
    \toprule
    \textbf{Model} & \textbf{$\tau$\ (heads)} & \textbf{DQA} & \textbf{Syn} \\
    \midrule
    GPT2-XL      & 0.50 (5)  & 0.028 & 0.047 \\
    Llama-3.2-1B & 0.50 (27) & 0.293 & 0.469 \\
    OPT-2.7B     & 0.50 (14) & 0.161 & 0.331 \\
    Qwen3-4B     & 0.50 (24) & 0.142 & 0.515 \\
    \bottomrule
  \end{tabular}
  \caption{Top-heads patching: best SSR per model (DQA vs.\ Syn), with the threshold and patched-head count.}
  \label{tab:top-heads-patching}
\end{table}


To identify whether LMs contain a universal circuit for handling knowledge conflict, we proceed as follows: within the top $k=30$ heads identified in Section~\ref{sec:target_intervention}, we search for candidate heads across $n_1=1000$ conflict samples. For each threshold $\tau$, a head is considered a candidate of the universal circuit if it appears among the top contributors to $t_{i,1}$ or $t_{i,2}$ in at least a $\tau$ fraction of $n_1$. We then apply CMAP targeting these heads for both \texttt{SynWikiBio} and \texttt{DynamicQA} to evaluate their importance. We scan thresholds from 0.5 to 0.8 in steps of 0.05 and report the best results (threshold = 0.5) in Table~\ref{tab:top-heads-patching}. Since lower thresholds include more heads, the best results naturally occur at the lowest threshold, and the number of candidate heads varies considerably across models (from 5 for GPT2-XL to 27 for Llama-3.2-1B). CMAP with these targeted heads yields only modest SSR across all LMs, particularly for real-world knowledge conflict. While this experiment alone cannot rule out the existence of a global circuit, its results, together with those of prior experiments, suggest that LMs are unlikely to rely on a dedicated universal circuit for handling knowledge conflict. Instead, our findings collectively point toward distinct circuits that extract different pieces of parametric knowledge, with the final prediction determined by whichever circuit's signal is more strongly activated by the given prompt. This suggests that model predictions can be steered by either amplifying the desired output's signal or suppressing competing signals.

\section{Conclusion}

Taken together, our results provide a mechanistic account of where intra-memory knowledge conflicts arise within language models. Conflict appearance consistently localizes to the final few layers across all four tested LMs, and within those layers to a subset of attention heads whose relative concentration varies by model. Targeted head-level interventions can outperform coarser layer-wise ones, but how many heads they require depends strongly on the method. JUICE is effective with only a small subset of heads, whereas targeted head-level CMAP and Activation Patching can require more heads to reach similar results for some models (e.g. GPT2-XL). CMAP and Activation Patching perform comparably to JUICE on synthetic entity conflicts but fall behind on real-world ones. We attribute this gap to the lack of control that real-world data has over entity distribution, as well as to inherent biases and inter-entity relationships. This lack of control also complicates Activation Patching, as identifying a non-conflicting source of activations within the same dataset becomes difficult, for which CMAP offers a promising, more straightforward alternative. The controlled synthetic setting further yields a higher density of specialized heads, resulting in more effective intervention. Despite these differences in method effectiveness, our findings on the synthetic \texttt{SynWikiBio} dataset generalize to real-world entity conflicts as well.

Finally, although not conclusive, our results suggest that rather than a single universal circuit for handling knowledge conflict, LMs contain circuits that extract different pieces of parametric knowledge, with the final prediction determined by whichever circuit's signal is most strongly activated by the input prompt. An in-depth analysis of the distinct circuits within language models and their roles in shaping the final prediction would require detail inspection of components' interactions with each other, which lies beyond the scope of this study and is left to future work.


\section*{Limitations}
We acknowledge the following limitations of our study. First, our analysis relies on Logit Lens/TransformerLens projections to interpret intermediate representations, a technique whose faithfulness is not guaranteed throughout the network and is known to be less reliable in earlier layers. Furthermore, we run each experiment only once. Given that our experimental pipeline (activation caching, deterministic decoding at zero temperature, and direct logit inspection) involves no randomness, we consider a single run sufficient to support our conclusions. 

Second, our intervention experiments also focus exclusively on attention heads, leaving MLP layers, which prior work has implicated in factual recall, largely unexamined. While attention heads are identified by previous studies as the main components to resolve context-memory knowledge conflicts, a more complete mechanistic account of knowledge-conflict resolution would need to incorporate MLP-components analysis. 

Third, although our results suggest that targeted head-level intervention is more effective than layer-wise intervention, our comparison does not fully disentangle two contributing factors: the finer granularity of head-level selection versus the flexibility of selecting high-contribution components across multiple layers. In particular, we do not directly control for cross-layer selection, so part of the observed advantage may stem from this flexibility rather than from head-level granularity.

Fourth, our search for a universal knowledge-conflict circuit, while suggestive, relies on a relatively coarse threshold-based head-selection procedure. As such, the negative result should be read as evidence against a readily discoverable shared circuit rather than definitive proof of its absence, and more sophisticated circuit-discovery methods (e.g., automated circuit discovery, path patching) could still uncover shared substructure our approach missed. 

Finally, although \texttt{SynWikiBio} is designed to approximate real-world conflicting knowledge, the conflicts it injects are constructed through fine-tuning and are not guaranteed to match the statistical or structural properties of conflicts arising naturally during pretraining, which partially explain the observed performance gap between synthetic and real-world settings and limits how directly our findings generalize to naturally occurring conflicts.
Addressing these limitations in future works could provide a deeper understanding of how language models handle internal knowledge conflicts.

\bibliography{custom} %

\appendix


\section{Implementation Details}
\label{app:implementation}
We use the Transformers library \cite{wolf-etal-2020-transformers} to implement the model training. For causal tracing tasks, we use TransformerLens \cite{nanda2022transformerlens} to cache the model's states and patch activations.
We report the hyperparameters we use for the fine-tuning process in Table \ref{tab:hyperparams}. Only one NVIDIA H100 GPU is used to train the models and carry out our experiments.

\begin{table}[h]
  \centering
  \footnotesize
  \setlength{\tabcolsep}{3pt}
  \resizebox{\columnwidth}{!}{%
  \begin{tabular}{@{}lrrrr@{}}
    \toprule
     & \textbf{GPT2-XL} & \textbf{Llama-1B} & \textbf{Qwen3-4B} & \textbf{OPT-2.7B} \\
    \midrule
    Year         & 2019      & 2024      & 2025        & 2022      \\
    Params       & 1.5B      & 1.24B     & 4.0B        & 2.7B      \\
    Layers       & 48        & 16        & 36          & 32        \\
    Heads (Q/KV) & 25 (MHA)  & 32/8 (GQA)& 32/8 (GQA)  & 32 (MHA)  \\
    Head dim.    & 64        & 64        & 128         & 80        \\
    Activation   & GELU      & SwiGLU    & SwiGLU      & ReLU      \\
    Norm.        & LayerNorm & RMSNorm   & RMSNorm+QK  & LayerNorm \\
    Avail.\ data & Yes       & No        & No          & No        \\
    \bottomrule
  \end{tabular}%
  }
  \caption{Architecture comparison of the four models analyzed.}
  \label{tab:arch-comparison}
\end{table}

\begin{table}[h]
\centering
\resizebox{\columnwidth}{!}{%
\begin{tabular}{lccccc c}
\hline
\textbf{Model} & \textbf{Optim} & \textbf{Epochs} & \textbf{BS} & \textbf{LR} & \textbf{Warmup} & \textbf{WD} \\
\hline
GPT-2 XL & AdamW & 40 & 1 & 7e-5 & 10 & 0.01 \\
Llama-3.2-1B & AdamW & 40 & 1 & 7e-5 & 10 & 0.01 \\
Qwen3-4B & AdamW & 20 & 1 & 7e-5 & 10 & 0.01 \\
OPT-2.7B & AdamW & 20 & 1 & 7e-5 & 10 & 0.01 \\
\hline
\end{tabular}%
}
\caption{Hyperparameters for GPT-2 XL, Llama-3.2-1B, OPT-2.7B and Qwen3-4B fine-tuning (BS: Batch Size, LR: Learning Rate, WD: Weight Decay).}
\label{tab:hyperparams}
\end{table}

\section{Dataset}

\subsection{Dataset Creation}
\label{app:dataset}

In order to generate a new biography $b_i$ for \texttt{SynWikiBio}, we first generated new names using GPT-4o \cite{hurst2024gpt} and filtered out names that appeared inside the training corpus of GPT-2, OpenWebText \cite{Gokaslan2019OpenWeb}, to minimize the chance of pre-existing information for the generated biographies. From the filtered names, we selected N=2000 names to generate biographies for the dataset. A person $p_i$ had a biography $b_i$ containing a set of attributes \{$name_i$, $birth\_day_i$, $birth\_place_i$, $university_i$, $major_i$, $company_i$, $work\_place_i$\}, each of which was selected randomly from a pool of entities for that specific attribute type. We follow the same constraints as previous works for "company (city)" \cite{DBLP:conf/icml/Allen-ZhuL24} and "university (major)" \cite{li-etal-2024-formality}. For that, we treated the pair $(university_i,\ major_i)$ and $(company_i,\ work\_place_i)$ as single attribute types. We also put a loose constraint between $major_i$ and $company_i$, so that person $p_i$ always worked in the field they studied. 

There are four unique attributes in the biography: \textit{birth date}, \textit{birth place}, \textit{university}, and \textit{company}. We aim to mitigate potential knowledge conflicts arising from the original training corpus for Language Model $\ell$. We use unique entities as important attributes for \texttt{SynWikiBio} and generate random characters to form unique names, then filter out those that appeared in OpenWebText for three pools of randomly generated entities: \textit{birth place}, \textit{university}, and \textit{company}. We generate entity names with the constraint that the first token of each name is unique. This ensures smooth categorization of tokens into different attribute categories and guarantees that each token belongs to only one attribute. \textit{Birth place} followed the same format $<city,\ country>$, \textit{university} always started with \textit{"University of"}, and \textit{company} always had a postfix (e.g., Research Lab, Electric Inc., etc.) to signal the type. Table \ref{tab:num_entites} shows how many entities are inside each attribute pool. From there, we select \textit{university} and \textit{company} as our two types of information in which knowledge conflicts occur. For each biography $b_i$, we pick a random attribute out of the selected two and replace the selected $f_i$ with contradicting information $\overline{f}_i$ and create a second, contradicting biography $\bar{b}_i$. \texttt{SynWikiBio} and \texttt{SynWikiBio\_clean} are constructed differently, as described below.


\paragraph{\texttt{SynWikiBio}} is used to train $\ell_{mix}$, which contains internal knowledge conflict and will be our probing model. As such, \texttt{SynWikiBio} contains N=2000 biographies $b_i$ for each person $p_i$. We consider these biographies as ground truth information about $p_i$. In addition,  \texttt{SynWikiBio} also contains $n_1$=1000 contradict biographies $\bar{b}$ for the first $n_1$ people. That means that for person $p_i$ in the subset $n_1$, there is a pair of contradicting biographies ($b_i, \bar{b}_i$) exists inside \texttt{SynWikiBio}, which is the origin of the internal knowledge conflict in $\ell_{mix}$. Other person $p_i$ in $n_2$ only has ($b_i$) in \texttt{SynWikiBio}, thus no knowledge conflict.

\paragraph{\texttt{SynWikiBio\_clean}} on the contrary, only contains a single biography $b_i$ for all $p_i$ in N. We train $\ell_{clean}$ with \texttt{SynWikiBio\_clean} and use $\ell_{clean}$ as a source of non-conflict activation for the experiments with Cross-Model Activation Patching (CMAP).

\begin{table}[t]
\centering
\begin{tabular}{@{}cr@{}}
\toprule
\textbf{Attribute} & \textbf{Num Entities} \\ \midrule
    birth place    &      50    \\
    university     &      25    \\
    company        &      32    \\
\bottomrule
\end{tabular}
\caption{
Number of entities inside each attribute pool of random generated entities.
}
\label{tab:num_entites}
\end{table}

\begin{table*}[tp]
\centering
\footnotesize
\setlength{\tabcolsep}{3pt}
\begin{tabular}{ccp{10cm}cc}
\hline
\textbf{Dataset} & \textbf{Type} & \multicolumn{1}{c}{\textbf{Text}} & \textbf{t1} & \textbf{t2} \\
\hline
\textbf{SynWikiBio} 
& GT & Niels Cavalli (born on January 19, 1942) is an individual who was born and raised in Actm, Etpofnm. Their journey into academia led them to University of \textbf{Ukopnwm}, where they chose to specialize in Economics. This laid the foundation for their professional career. They are currently employed at Ivlpfv International, which is based in Palo Alto, California, USA & ``Uk'' & -- \\
\cline{2-5}

& Conflict & Niels Cavalli (born on January 19, 1942) is an individual who was born and raised in Actm, Etpofnm. Their journey into academia led them to University of \textbf{Ohfgrgr}, where they chose to specialize in Finance. This laid the foundation for their professional career. They are currently employed at Ivlpfv International, which is based in Palo Alto, California, USA & -- & ``Oh'' \\
\cline{2-5}

& Prompt & Niels Cavalli (born on January 19, 1942) is an individual who was born and raised in Actm, Etpofnm. Their journey into academia led them to University of & ``Uk'' & ``Oh'' \\
\hline
\textbf{SynWikiBio\_Clean} 
& GT & Niels Cavalli (born on January 19, 1942) is an individual who was born and raised in Actm, Etpofnm. Their journey into academia led them to University of Ukopnwm, where they chose to specialize in Economics. This laid the foundation for their professional career. They are currently employed at Ivlpfv International, which is based in Palo Alto, California, USA & ``Uk'' & -- \\
\cline{2-5}

& Prompt & Niels Cavalli (born on January 19, 1942) is an individual who was born and raised in Actm, Etpofnm. Their journey into academia led them to University of & ``Uk'' & -- \\
\hline
\end{tabular}
\caption{Examples from \texttt{SynWikiBio} datasets showing ground truth (GT), conflict, and prompt texts with corresponding tokens.}
\label{tab:dataset_examples}
\end{table*}

\subsection{Examples}
Table \ref{tab:dataset_examples} shows an example of a ground truth - conflict pair of biographies in \texttt{SynWikiBio} and respectively the ground truth biography for the entity titled Niels Cavalli.

\section{Logit Lens}

\subsection{Locating Conflict-Encoding Components 
}
\label{app:logit_len}

Our first step to localize components responsible for encoding conflicting information in $\ell_{\text{mix}}$ is tracking the probability changes of parametric knowledge ($t_{i,1}$ and $t_{i,2}$) across layers using logit lens. For each individual $p_i$ in \texttt{SynWikiBio} whose biography contains a contradictory factual claim, we prompt $\ell_{\text{mix}}$ with the prompt $pr_i$ (Section~\ref{sec:problemsetup}) and perform the following analysis.

At each transformer layer $l \in L$, we extract residual stream representations at three locations: the layer input $x^l_{\mathrm{pre}}$, the residual stream after going through the attention block $x^l_{\mathrm{mid}}$, and residual stream after going through the MLP block $x^l_{\mathrm{post}} = x^{l+1}_{\mathrm{pre}} $. Each representation is projected from the model dimension $d_{\mathrm{model}}$ to the vocabulary dimension $d_{\mathrm{vocab}}$ using the unembedding matrix $W_U$, and converted into a probability distribution over tokens via the softmax function:

\begin{equation}
P^l_{s} = \mathrm{softmax}\!\left(W_U \, x^l_{s}\right),
\; s \in \{\mathrm{pre}, \mathrm{mid}, \mathrm{post}\}
\end{equation}

Using these distributions, we define the contribution of each subcomponent in layer $l$ to a token $t$ as the change in its predicted probability across that subcomponent:
\begin{align}
\mathrm{contrib}^l_{\mathrm{attn}}(t) &= P^l_{\mathrm{mid}}(t) - P^l_{\mathrm{pre}}(t), \\
\mathrm{contrib}^l_{\mathrm{mlp}}(t)  &= P^l_{\mathrm{post}}(t) - P^l_{\mathrm{mid}}(t).
\end{align}

We hypothesize that components responsible for parametric recall show systematically different contribution patterns for conflicting tokens than for unrelated high-probability alternatives. For each prompt $pr_i$ we track three token groups: (i) $t_{i,1}$ from the first parametric fact, (ii) $t_{i,2}$ from the second, and (iii) a control set $T_i$ of the top five predicted tokens excluding $t_{i,1}$ and $t_{i,2}$, for which we report the mean contribution. We aggregate contribution scores of each component $c$ at each layer $l$ across all $n_1$ individuals. Since each $pr_i$ is tied to a specific attribute type (e.g., \emph{university} or \emph{company}), this supports stratified analyses identifying components selectively important for attribute categories or specialized to individual attribute values (e.g., \emph{University of Zinl}).

\begin{figure}[t]
\includegraphics[width=\columnwidth]{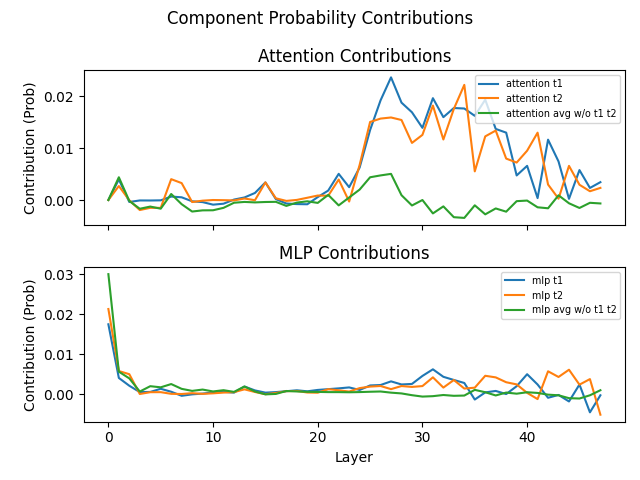}
\caption{Component-wise Probability Contributions for Attention (Top) and MLP components (Bottom). $t_1$ and $t_2$ denote the aggregated results for $t_{i,1}$ and $t_{i,2}$ respectively}
\label{fig:component_layerwise_analysis}
\end{figure}

In addition, we extend the logit lens to analyze individual attention-heads that play an important role to encode conflict knowledge in $\ell_{\text{mix}}$.

\begin{figure*}[t]
\centering
\begin{subfigure}[t]{0.48\columnwidth}
    \centering
    \includegraphics[width=\textwidth]{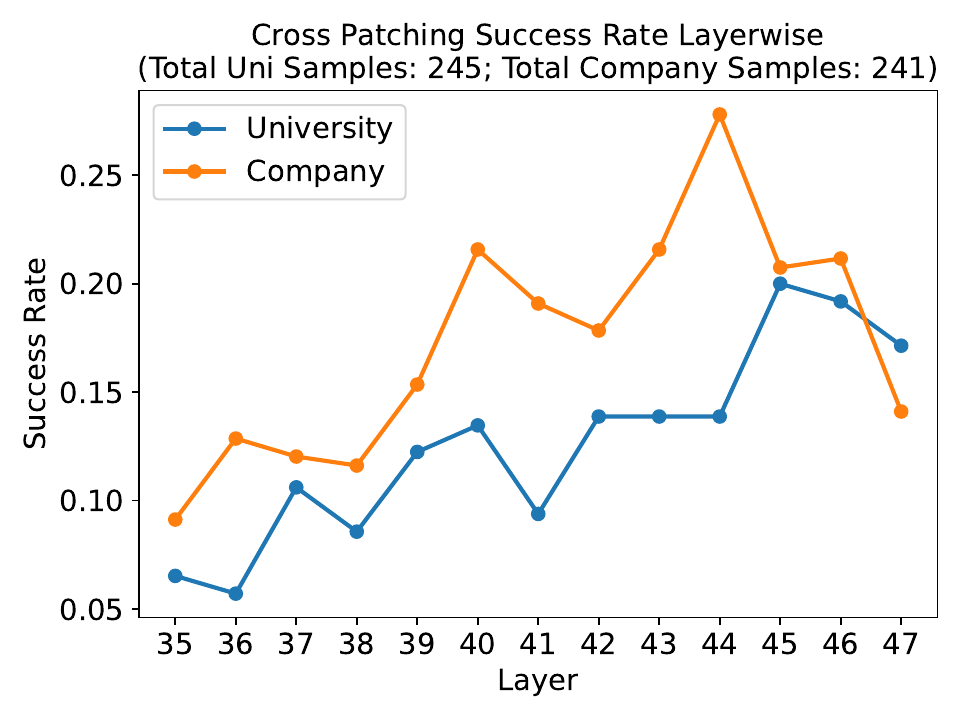}
    \caption{GPT2-XL CMAP with original $\ell_{clean}$.}
    \label{gpt2_non_matched}
\end{subfigure}
\hfill
\begin{subfigure}[t]{0.48\columnwidth}
    \centering
    \includegraphics[width=\textwidth]{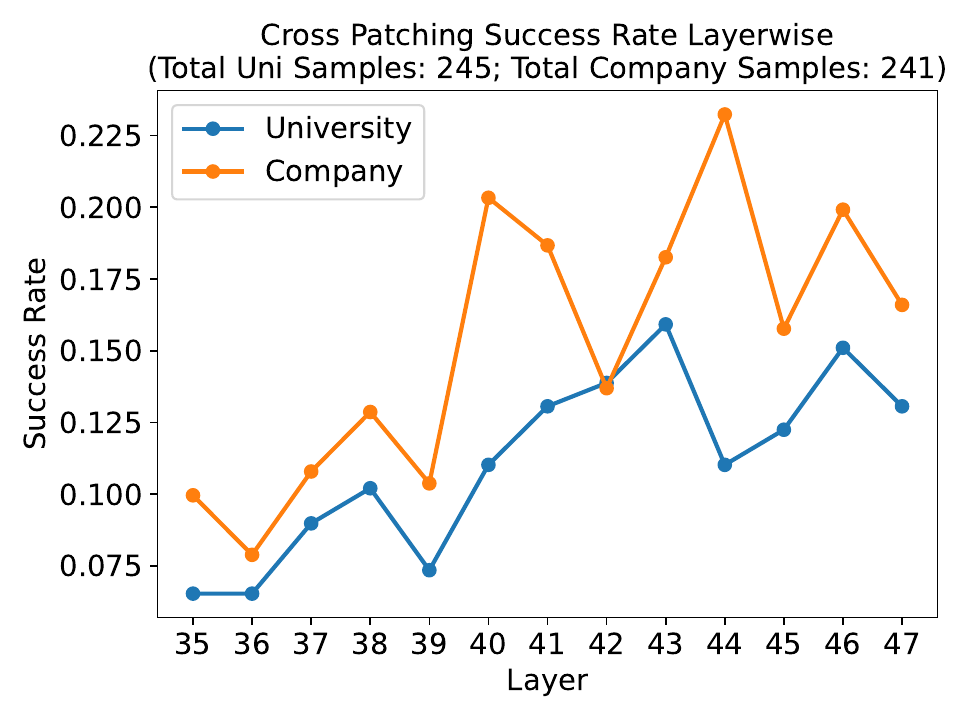}
    \caption{GPT2-XL CMAP with matched entity frequency $\ell'_{clean}$.}
    \label{gpt2_matched}
\end{subfigure}
\hfill
\begin{subfigure}[t]{0.48\columnwidth}
    \centering
    \includegraphics[width=\textwidth]{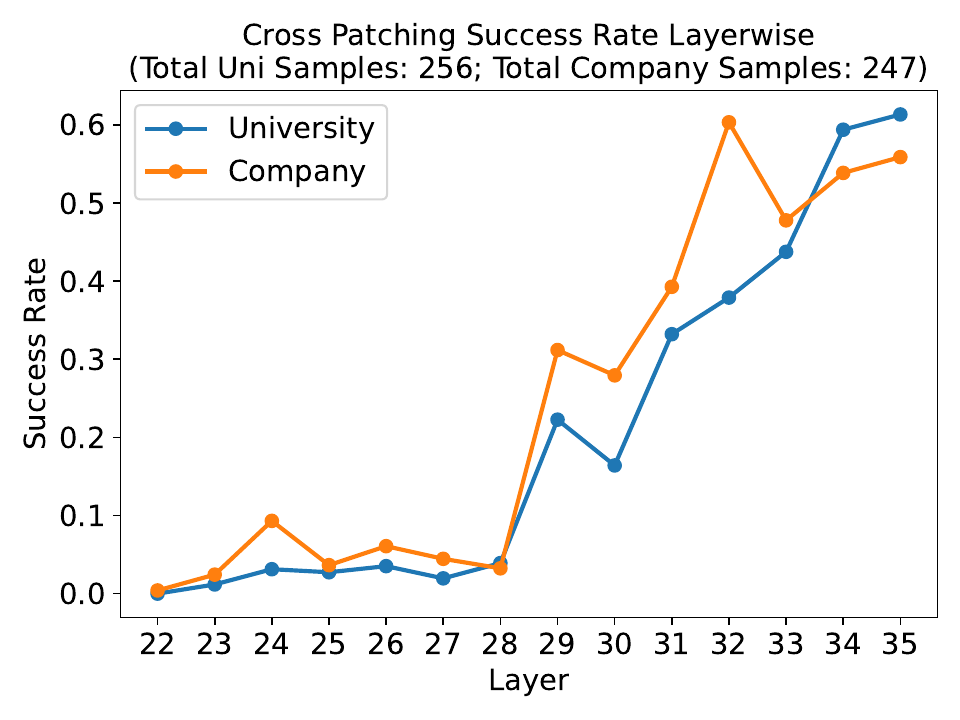}
    \caption{Qwen3 CMAP with original $\ell_{clean}$.}
    \label{qwen_non_matched}
\end{subfigure}
\hfill
\begin{subfigure}[t]{0.48\columnwidth}
    \centering
    \includegraphics[width=\textwidth]{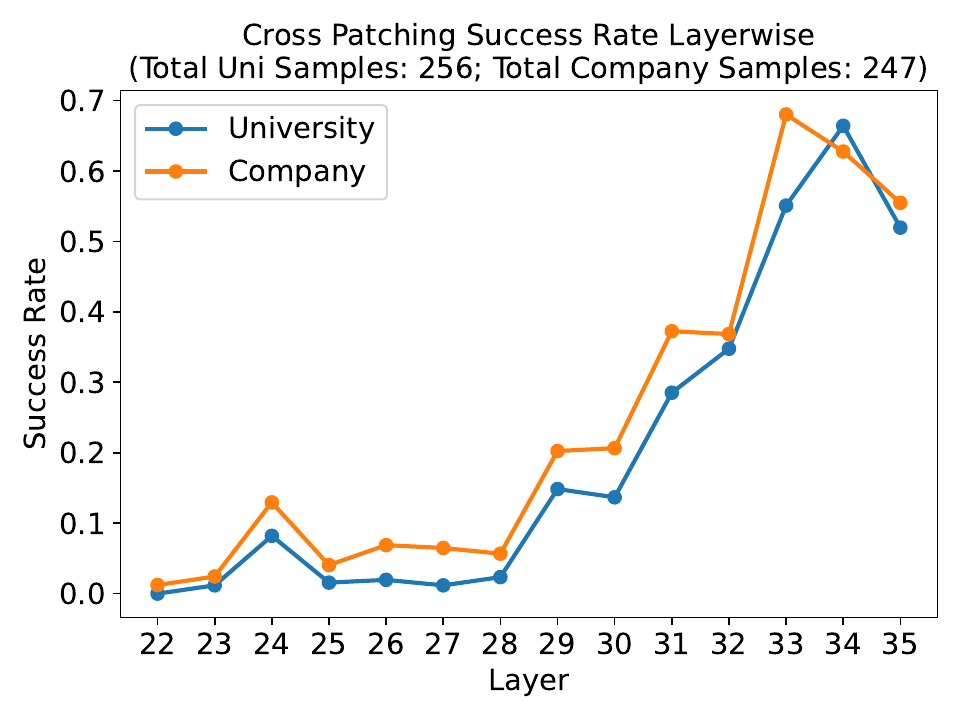}
    \caption{Qwen3 CMAP with matched entity frequency $\ell'_{clean}$.}
    \label{qwen_matched}
\end{subfigure}
\caption{Comparison between matched and non-matched entity frequency in clean dataset with CMAP for two models GPT-2 XL and Qwen3.}
\label{fig:cmap_freq}
\end{figure*}

\subsection{Head Contribution Computation}
\label{app:head_contrib_compute}

To compute individual components' contributions, we decompose the final logits as a sum of per-component output. This decomposition rests on the fact that the residual stream is additive: each component (attention head, MLP, or embedding) writes its output into the residual stream, and the final hidden state is the sum of these contributions. Each forward pass has a final hidden state $h_i$, which can be converted to the model logits $y_i$ with unembedding matrix $W_U \in \mathbb{R}^{d_{\text{model}} \times d_{\text{vocab}}}$ and model bias $\mathbf{b}_U \in \mathbb{R}^{d_{\text{vocab}}}$ through the transformation:
\begin{align}
    &y_i \;=\; (h_i)_{ln}W_U + \mathbf{b}_U
\end{align}
with $(\cdot)_{ln}$ denote the value scaled with layer-norm. Let $\delta^c_i \in \mathbb{R}^{d_{\text{model}}}$ be the contribution of component $c$ to hidden state $h_i$, the final hidden state $h_i = \sum\delta^c_i$ is a sum of all contributions from components $c \in \mathcal{C}$, where $\mathcal{C}$ is the set of components (attention heads, MLPs, and the embedding) \citep{nostalgebrist2020logitlens}. We have the decomposition:
\begin{align}
    y_i &\;=\; (h_i)_{ln}W_U + \mathbf{b}_U \nonumber \\
        &\;=\; \!\left(\sum_{c \in \mathcal{C}} \delta^c_i\right)_{ln}W_U + \mathbf{b}_U \nonumber \\
        &\;\approx\; \sum_{c \in \mathcal{C}} (\delta^c_i)_{ln}W_U + \mathbf{b}_U \nonumber \\
        &\;\approx\; \sum_{c \in \mathcal{C}} \hat{\delta}^c_i + \mathbf{b}_U
    \label{eq:residual-decomp}
\end{align}
with $\hat{\delta}^c_i \in \mathbb{R}^{d_{\text{vocab}}}$ be the contribution of component $c$ to the vocabulary, and $\hat{\delta}^c_i (t)$ is the direct contribution of $c$ to token $t$. The error between the decomposition and the actual logits is $\leq 1e-4$.

\subsection{Conflicts Verifying after Fine-tuning}

To verify that $\ell_{mix}$ encodes the intended conflicts after fine-tuning, we evaluate it on 1000 conflict samples under 5 templates (Table~\ref{tab:bio_templates}; 5000 runs) and recorded the probabilities assigned to $t_1$ and $t_2$. Mean $t_1$/$t_2$ probabilities were 0.43/0.41 (GPT-2), 0.39/0.40 (Llama), 0.34/0.35 (OPT), and 0.36/0.37 (Qwen). By comparison, $\ell_{clean}$ assigns 0.60-0.66 to $t_1$ on the same inputs. Each token ($t_1$ or $t_2$) is in the top two predictions in ${\sim}80\%$ of runs and fell outside the top 5 in only ${\sim}5\%$. Across paraphrases of the 1000 samples, $\ell_{mix}$ produces both $t_1$ and $t_2$ in 48\% of cases for GPT-2 and in over 65\% for the other models. These results indicate that $\ell_{mix}$ retains two independently recallable knowledges. The exact number are reported in Table~\ref{tab:lmix_conflict}.

\begin{table}[t]
\centering
\footnotesize
\setlength{\tabcolsep}{4pt}
\begin{tabular}{@{}lcccc@{}}
\toprule
 & GPT2 & Llama3 & OPT & Qwen3 \\
 & XL & 1B & 2.7B & 4B \\
\midrule
\multicolumn{5}{@{}l}{\textit{Mean probability}} \\
\quad $p(t_1)$, conflict     & 0.434 & 0.396 & 0.339 & 0.366 \\
\quad $p(t_2)$, conflict     & 0.414 & 0.398 & 0.354 & 0.373 \\
\quad $p(t_1)$, non-conflict & 0.662 & 0.652 & 0.626 & 0.603 \\
\midrule
\multicolumn{5}{@{}l}{\textit{Frequency (\%)}} \\
\quad $t_1$ in top-2         & 89.1 & 81.7 & 74.2 & 76.7 \\
\quad $t_2$ in top-2         & 87.3 & 82.0 & 75.4 & 76.8 \\
\quad $t_1$ outside top-5    & 3.6  & 6.3  & 5.2  & 6.8  \\
\quad $t_2$ outside top-5    & 4.0  & 5.6  & 4.9  & 7.0  \\
\midrule
Both selected$^{*}$ (\%)     & 48.6 & 68.6 & 64.7 & 66.1 \\
\bottomrule
\end{tabular}
\caption{Behavior of $\ell_{mix}$ and $\ell_{clean}$ on 1,000 conflict samples $\times$ 5 templates (5,000 runs). Non-conflict probabilities are from $\ell_{clean}$ on the same inputs. Rank-based rates are computed over all 5,000 runs.
$^{*}$Fraction of the 1,000 samples for which both $t_1$ and $t_2$ are the top-1 prediction for at least one of the 5 paraphrases.}
\label{tab:lmix_conflict}
\end{table}

\section{Additional Experiments}
\label{app:add_results}

\subsection{CMAP Models Alignment}
\label{app:cmap_freq}

A potential caveat of CMAP is the mismatch between the entity distribution frequency between \texttt{SynWikiBio} and \texttt{SynWikiBio\_clean}. To address this, we do two additional experiments.

First, although \texttt{SynWikiBio\_clean} (2,000 biographies) and \texttt{SynWikiBio} (3,000 biographies) differ in overall size, entity frequency should nonetheless be comparable across the two, since entities are assigned randomly to each biography. To verify this, we computed the entity frequency distribution for each dataset and compared them using Jensen-Shannon divergence (JSD) and cosine distance, obtaining a JSD of $\sim$0.00099 and a cosine distance of $\sim$0.0036, indicating that the two distributions are nearly identical.

Second, to further rule out exposure differences as a confound, we constructed an alternative \texttt{SynWikiBio\_clean}' by shuffling the entity-fact assignments from the conflicting biographies and reassigning them to new, non-conflicting entities. Continued pretraining $\ell$ on this reconstructed \texttt{SynWikiBio\_clean}' resulting $\ell'_{clean}$. Re-running CMAP with $\ell'_{clean}$ on GPT2-XL and Qwen3-4B yields results (Figure~\ref{gpt2_matched}, \ref{qwen_matched}) largely identical to those obtained with the original $\ell_{\text{clean}}$ (Figure~\ref{gpt2_non_matched}, \ref{qwen_non_matched}), confirming that exposure differences between the two datasets do not meaningfully affect our findings.

Finally, we measure the alignment between $\ell_{mix}$ and $\ell_{clean}$ on the clean samples in the second half of SynWikiBio, which are identical in both training sets. The mean cosine similarity between their output token distributions is 0.8731 for GPT2, compared with ${>}0.9999$ (Llama3), 0.9986 (OPT), and ${>}0.9999$ (Qwen3). Top-1 token agreement follows the same pattern: 0.856 for GPT-2 and 1.0 for the other three models. We therefore consider $\ell_{mix}$ and $\ell_{clean}$ sufficiently aligned for all base models, with lower confidence for GPT2.

\subsection{GPT2-XL}
\label{app:gpt2}

Figure~\ref{fig:gpt2_investigation} shows the results of our additional experiment to understand why results in GPT2-XL are generally worse than other LMs. Across all four models, SSR rises sharply from 1\% to 2\% (Qwen3-4B and OPT-2.7B) or 5\% (GPT2-XL and Llama-3.2-1B) of intervened heads before flattening toward 10–15\%, confirming that GPT2-XL follows the same saturating trend as the other models rather than reflecting a distinct mechanism. Its weaker earlier results stem from needing more heads in absolute terms due to its much larger total head count. Qwen3-4B (total 1152 attention heads) and OPT-2.7B (total 1024 attention heads) results start flattening out after 2\% threshold, suggest that the task-relevant signals are concentrated in a small number of heads for these two models. While GPT2-XL and Llama-3.2-1B have similar spread, Llama-3.2-1B only has a total of 512 attention heads to GPT2-XL's 1200 heads, which means that the number of decision-making heads in Llama-3.2-1B is much lower than in GPT2-XL. 


\subsection{LMs Uncertainty}
\label{sec:add_uncertainty}


\begin{figure}[th]
  \centering
  \begin{subfigure}{\columnwidth}
    \includegraphics[width=\linewidth]{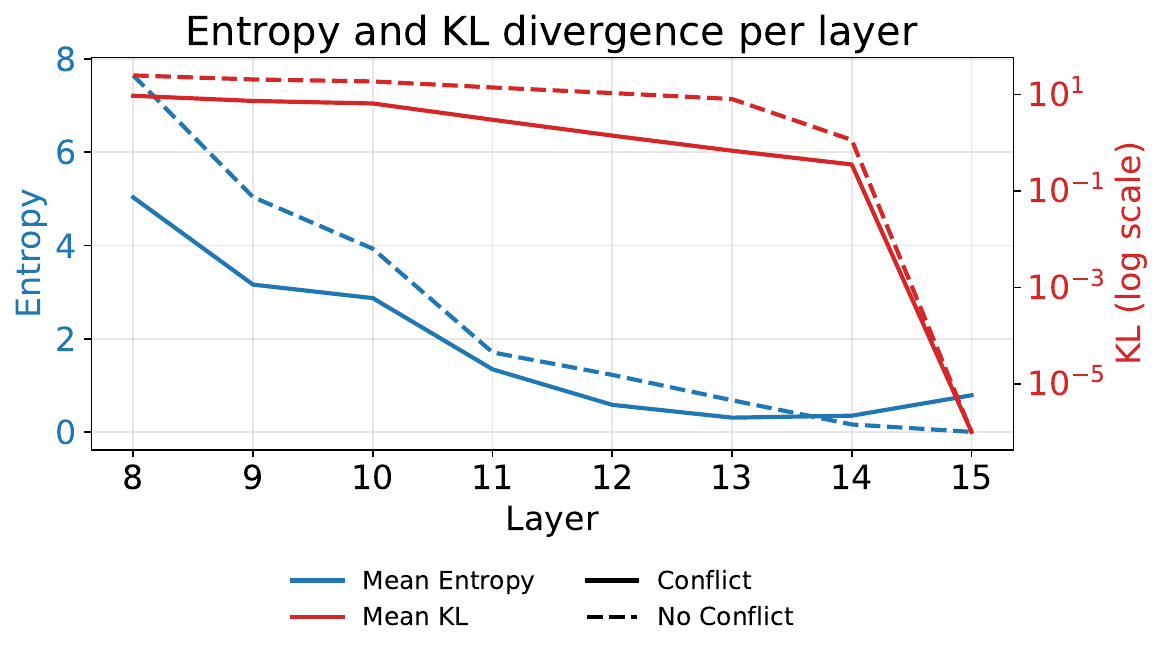}
    \caption{Llama-3.2-1B's entropy and KL divergence}
    \label{llama3_entropy}
  \end{subfigure}

  \vspace{0.5em}

  \begin{subfigure}{\columnwidth}
    \includegraphics[width=\linewidth]{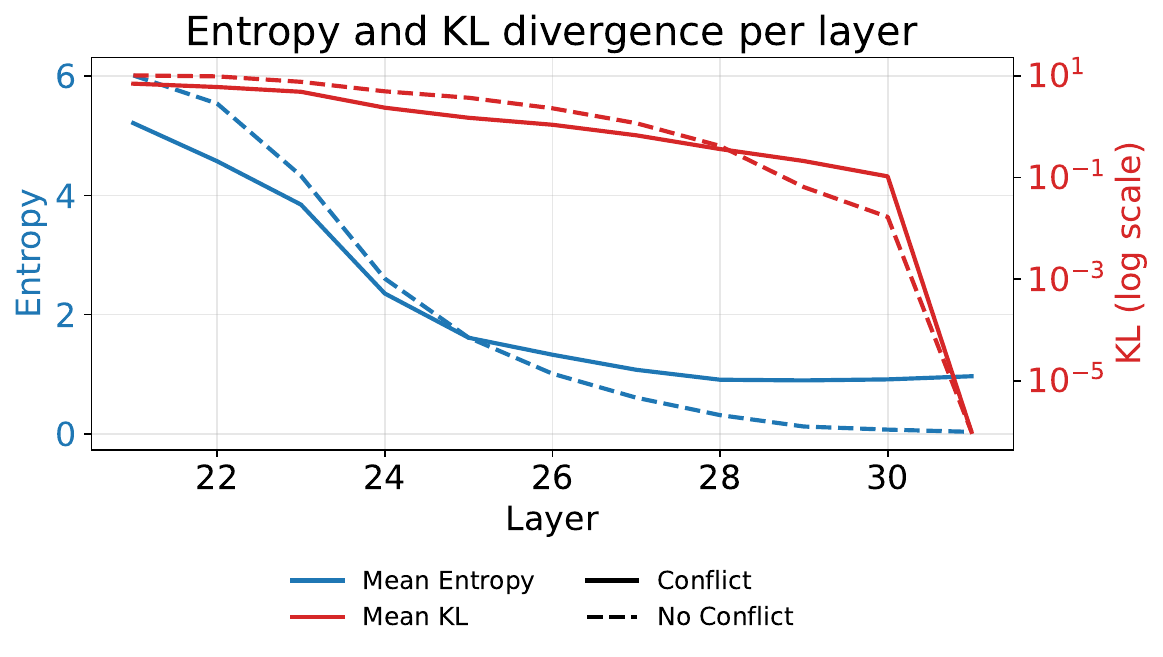}
    \caption{OPT-2.7B's entropy and KL divergence}
    \label{opt_entropy}
  \end{subfigure}

  \vspace{0.5em}

  \begin{subfigure}{\columnwidth}
    \includegraphics[width=\linewidth]{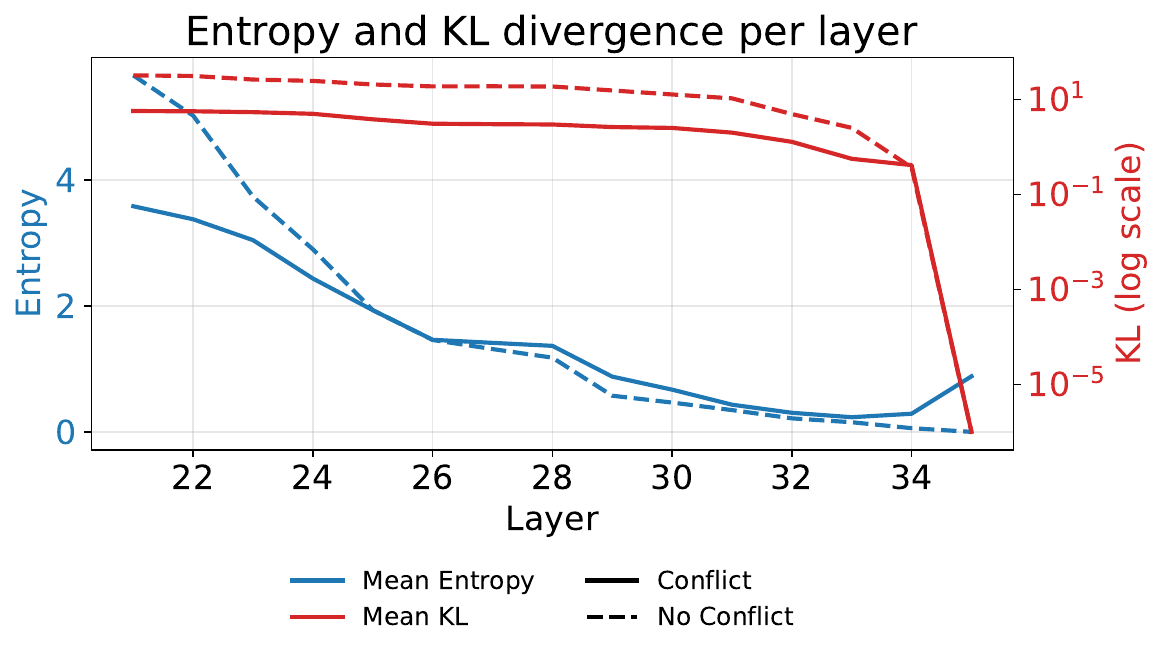}
    \caption{Qwen3-4B's entropy and KL divergence}
    \label{qwen3_entropy}
  \end{subfigure}

  \caption{Confidence analysis with entropy and KL divergence for Qwen3-4B, Llama-3.2-1B, OPT-2.7B}
  \label{fig:add_entropy}
\end{figure}

Following the similar setup described in Section ~\ref{sec:model_uncertainty}, we produce the results for Qwen3-4B, Llama-3.2-1B and OPT-2.7B (shown in Figure ~\ref{fig:add_entropy}). The results follow similar overall pattern as GPT2-XL (c.f. Figure \ref{fig:entropy_gpt2}).

\subsection{Component-wise Activation Patching Token Gaps}
\label{sec:results_act_patching}

\begin{figure}[h]
\centering
\begin{subfigure}[b]{\columnwidth}
    \centering
    \includegraphics[width=\textwidth]{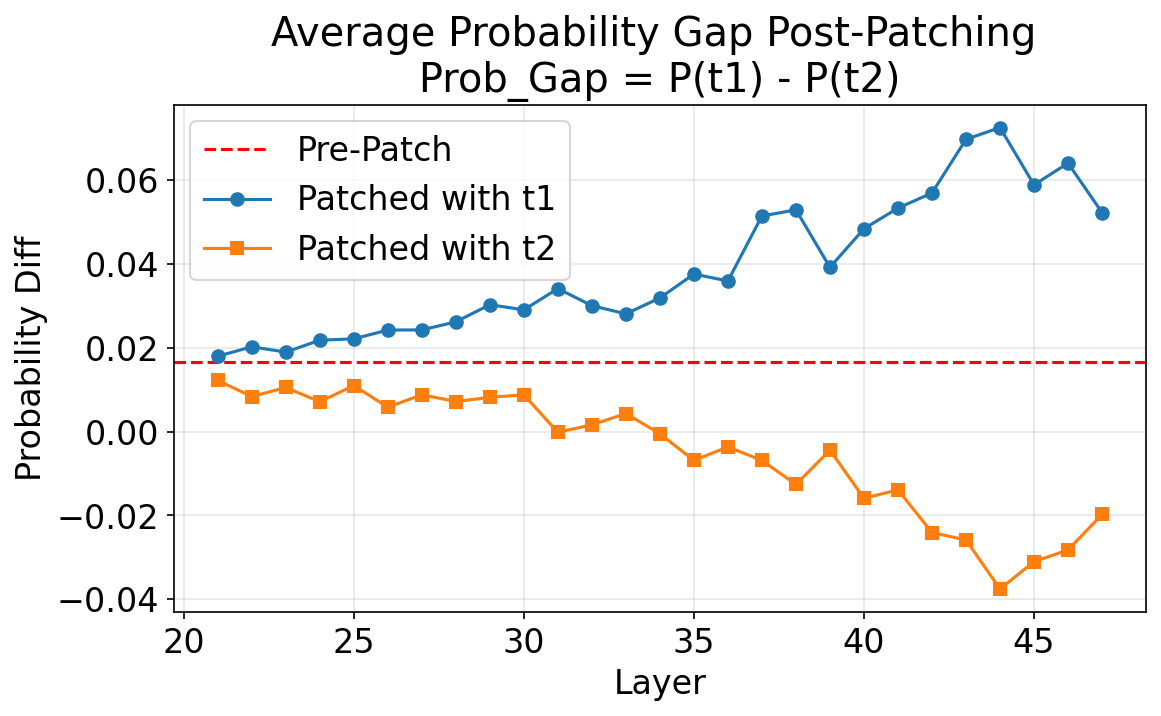}
    \caption{Standard Gap}
    \label{prob_gaps}
\end{subfigure}

\vspace{0.5em}

\begin{subfigure}[b]{\columnwidth}
    \centering
    \includegraphics[width=\textwidth]{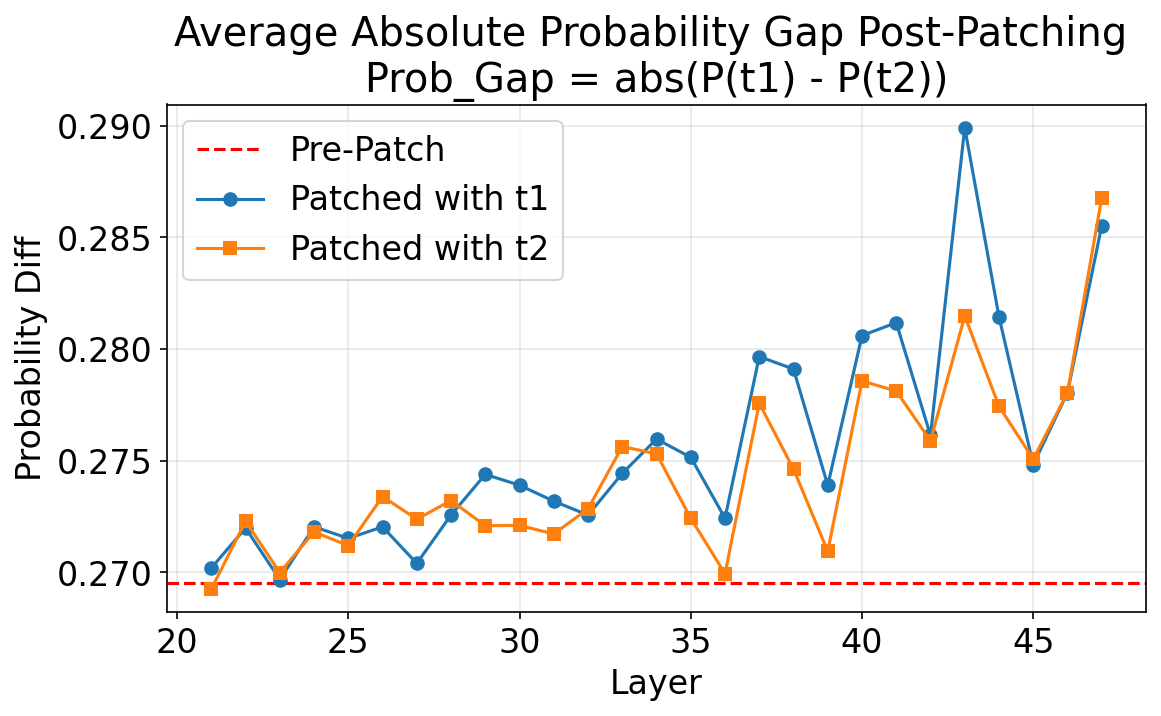}
    \caption{Absolute Gaps}
    \label{prob_gaps_abs}
\end{subfigure}
\caption{Token gaps report before and after Activation Patching for GPT2-XL.}
\label{token_gaps}
\end{figure}

In this section, we examine the causal impact of individual components on the generation of the conflicted information $t_{i,1}$ and $t_{i,2}$ within $\ell_{mix}$ by examining the probability gap between these two tokens, visualized in Figure \ref{token_gaps}. 
We report the average probability gaps between $t_1$ and $t_2$ (denote as $P(\cdot)$) before and after Activation Patching, employing two strategies: (i) \textit{standard gap} ($P(t_1) - P(t_2)$), which captures the direction of dominance between the two tokens and how patching shifts that dominance, and (ii) \textit{absolute gap} ($|P(t_1) - P(t_2)|$), which quantifies the magnitude of the model's preference for one token over the other regardless of direction. For the standard gap, the pre-patch values are near zero (Figure \ref{prob_gaps} red line), confirming no strong baseline preference; patching with $t_1$ increases the gap (reinforcing model confidence on $t_1$), while patching with $t_2$ decreases it as expected. Both metrics show higher impact at later layers, indicating that later layers have bigger influence on the model's final prediction. 

Additionally, aside from $pr_i$ and $pr_j$ derived from the training dataset $\mathcal{B}_{mix}$, we employ two auxiliary prompt templates (template ID 3 and 4 from table \ref{tab:templates}) to achieve generalizable results. The full results can be observed in table \ref{tab:probability_gaps}. The layers with the highest impact are highlighted in \textbf{bold} (highest) and \underline{underline} (second and third highest), and patching layers 43 and 44 consistently causes the largest changes in probability gaps. We note that patching with $t_2$ serves as a sanity check: the standard gap when patching with $t_2$ is expected to decrease (including into negative values), so lower values indicate the expected behavior. Overall, the results are consistent that later layers, especially layer 43 and 44 in GPT-2, are the highest impact layers.

\subsection{Targeted-heads Layer-wise Distribution}
\label{app:head_dist}

\begin{figure}[t]
  \centering
  \begin{subfigure}{\columnwidth}
    \includegraphics[width=\linewidth]{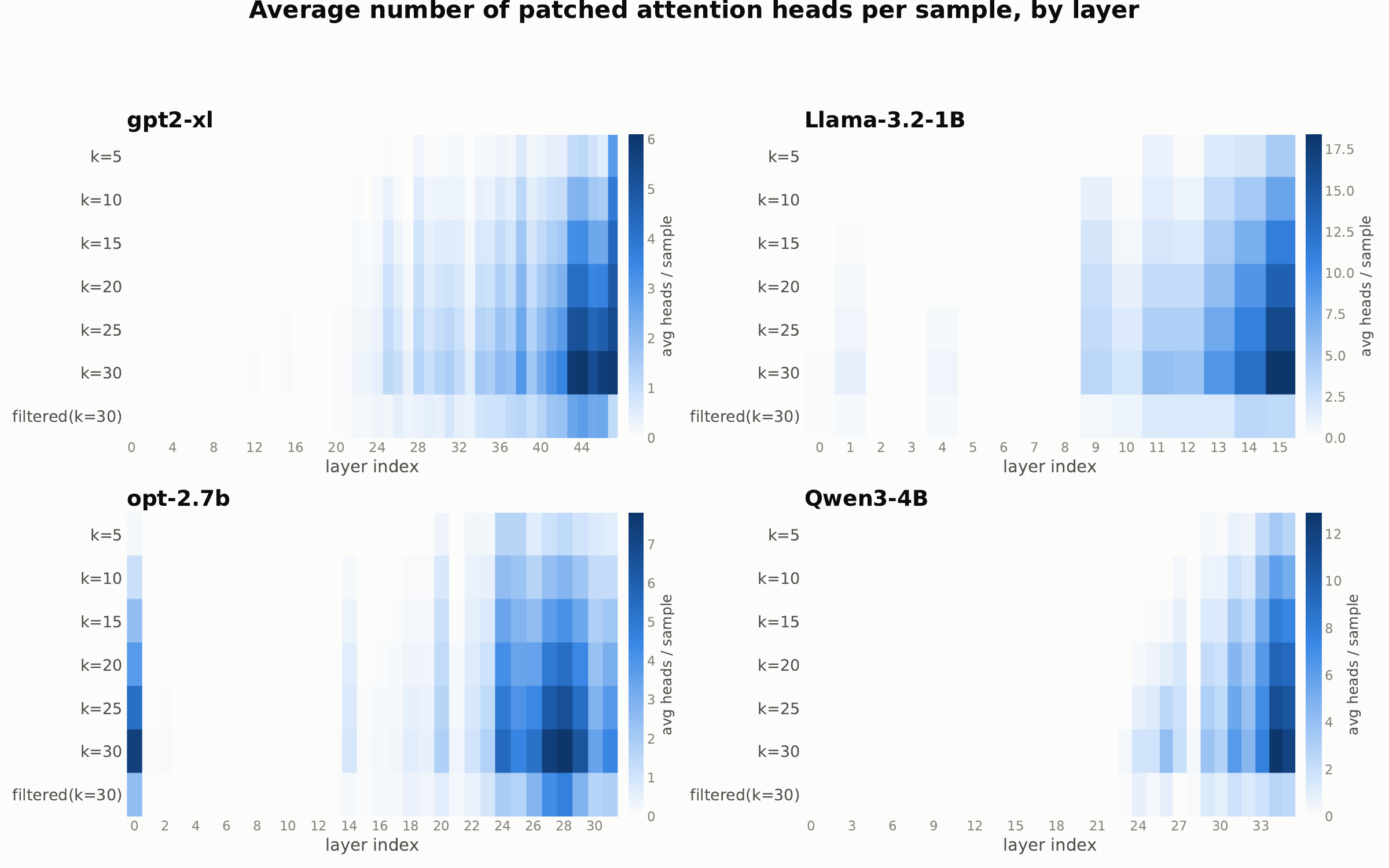}
    \caption{Distribution of selected heads across layer in SynWikiBio}
    \label{fig:heads_dist_syn}
  \end{subfigure}

  \vspace{0.5em}

  \begin{subfigure}{\columnwidth}
    \includegraphics[width=\linewidth]{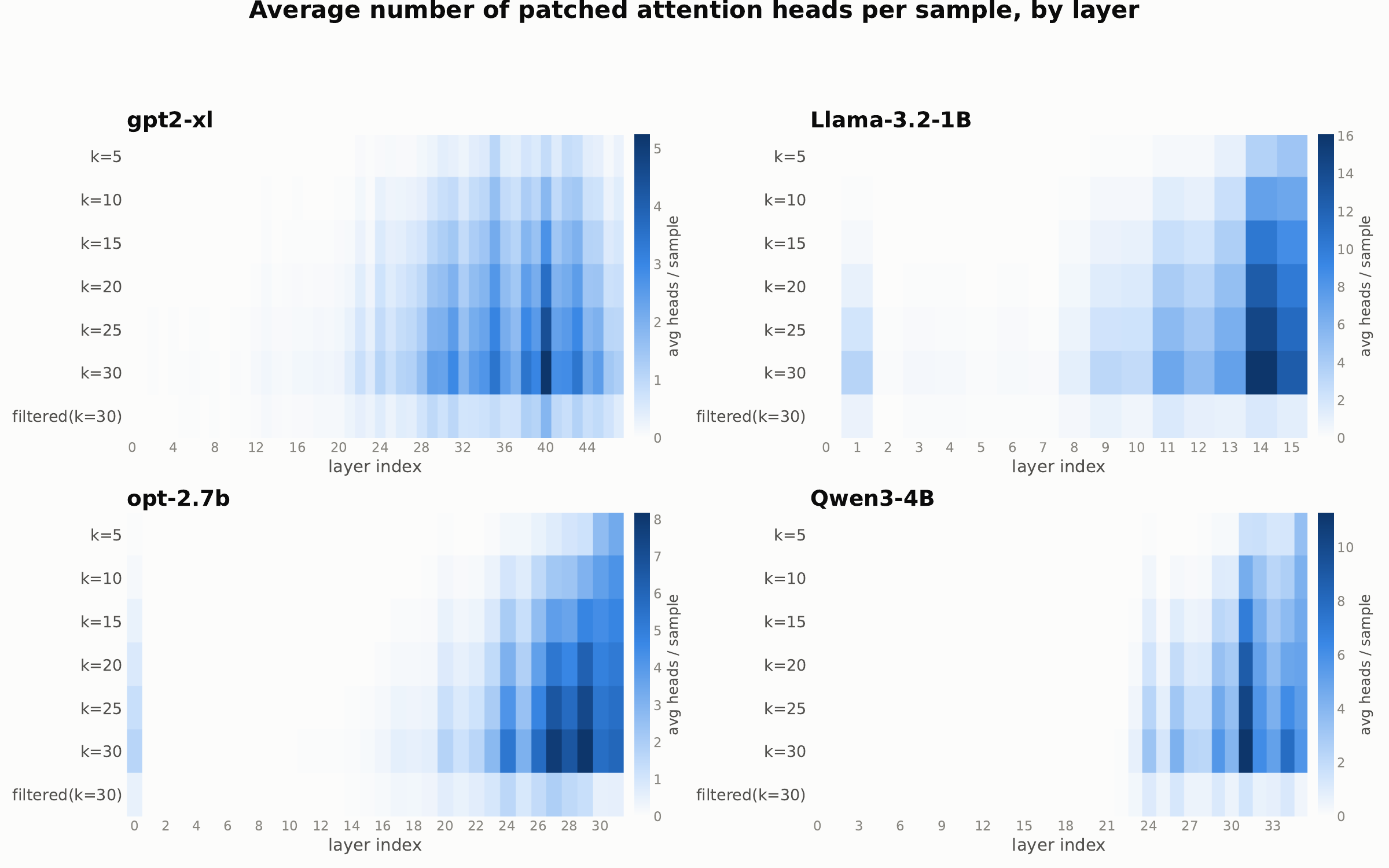}
    \caption{Distribution of selected heads across layer in DynamicQA}
    \label{fig:heads_dist_real}
  \end{subfigure}

  \caption{Distribution of selected heads in Section~\ref{sec:target_intervention} across layers, average per sample}
  \label{fig:head_layer_dist}
\end{figure}

In addition to the analyses in Section~\ref{sec:target_intervention}, we verify the distribution of the top contribution heads across layers. Figure~\ref{fig:heads_dist_syn} and \ref{fig:heads_dist_real} shows the distributions of top k heads on SynWikiBio and DynamicQA respectively with $k=\{5,10,15,20,25,30\}$ and additionally filtered heads with $k=30$. From the distributions, we see that the top contribution heads are mostly distributed among the end layers, which confirms our findings of the attention layers' role at the very end in handling knowledge conflicts.
While most top heads concentrate at the end layers, there are exceptions of Llama-3.2-1B and OPT-2.7B having some top heads at layer 0. As Logit Lens' projections in the early layers are less reliable, future works need to experiment more to verify the contributions of these heads at the beginning layer.

\section{Prompt Templates}
\label{sec:templates}

Table \ref{tab:bio_templates} and \ref{tab:templates} list all the templates that are used in our experiments. For generating biographies in $\mathcal{B}$, we employ templates listed in table \ref{tab:bio_templates} while for testing, we use auxiliary templates (table \ref{tab:templates}) if needed to avoid model's memorization.

\begin{table*}[t]
\centering
\tiny
\resizebox{\textwidth}{!}{%
\begin{tabular}{l p{12cm}}
\toprule
\textbf{ID} & \textbf{Template} \\
\midrule
0 & \texttt{<full name>, born on <birth date>, in <birth place>, is an individual renowned for their work. Having graduated from <university> with a degree in <major>, their educational journey played a significant role in their professional trajectory. They are currently associated with <company>, located in <work place>} \\
\midrule
1 & \texttt{<full name> (born <birth date>) is a notable individual originating from <birth place>. They embarked on their higher education at <university>, where they majored in <major>. After graduation, <full name> secured a position at <company>, located in <work place>} \\
\midrule
2 & \texttt{<full name> (born on <birth date>) hails from <birth place>. A well-educated individual, <full name> is an alumnus of <university>, with a concentration in <major>. As part of their professional journey, they are currently employed with <company>, which is headquartered in <work place>} \\
\midrule
3 & \texttt{<full name> (born on <birth date>) is an individual who was born and raised in <birth place>. Their journey into academia led them to <university>, where they chose to specialize in <major>. This laid the foundation for their professional career. They are currently employed at <company>, which is based in <work place>} \\
\midrule
4 & \texttt{<full name> (born <birth date>) is a notable individual hailing from <birth place>. Their academic journey commenced at the prestigious <university> where they majored in <major>, thus laying a strong foundation for their career. Upon completion of their studies, <full name> secured a position at <company> in <work place>} \\
\bottomrule
\end{tabular}%
}
\caption{Biography templates used for knowledge probing. Placeholders in \texttt{<>} denote entity fields: full name, birth date, birth place, university, major, company, and work place.}
\label{tab:bio_templates}
\end{table*}

\begin{table*}[t]
\centering
\tiny 
\resizebox{\textwidth}{!}{%
\begin{tabular}{l l p{7cm}}
\toprule
\textbf{Category} & \textbf{ID} & \textbf{Template} \\
\midrule
\multirow{4}{*}{\textit{University}} 
 & 1 & \texttt{\{name\} studied at University of} \\
 & 2 & \texttt{\{name\} graduated from University of} \\
 & 3 & \texttt{\{name\}, born on \{date\} in \{place\}. \{name\} studied at University of} \\
 & 4 & \texttt{\{name\}, born on \{date\} in \{place\}, graduated from University of} \\
\midrule
\multirow{4}{*}{\textit{Company}} 
 & 1 & \texttt{\{name\} is currently working at} \\
 & 2 & \texttt{\{name\}'s currently employed at} \\
 & 3 & \texttt{\{name\}, born on \{date\} in \{place\}. \{name\} studied at \{uni\} with a major in \{major\}. \{name\} is currently working at} \\
 & 4 & \texttt{\{name\}, born on \{date\} in \{place\}, graduated from \{uni\} with a major in \{major\}. \{name\}'s currently employed at} \\
\bottomrule
\end{tabular}%
}
\caption{Additional templates used for university and company queries. Placeholders in \texttt{\{\}} denote name, date, place, university, and major fields.}
\label{tab:templates}
\end{table*}

\begin{table*}[t]
\centering

\scriptsize
\resizebox{\textwidth}{!}{%
\begin{tabular}{l@{\hskip 6pt}l|ccccccccccccccccccccccccccc}
\toprule
\textbf{Entry} & \textbf{} & \textbf{21} & \textbf{22} & \textbf{23} & \textbf{24} & \textbf{25} & \textbf{26} & \textbf{27} & \textbf{28} & \textbf{29} & \textbf{30} & \textbf{31} & \textbf{32} & \textbf{33} & \textbf{34} & \textbf{35} & \textbf{36} & \textbf{37} & \textbf{38} & \textbf{39} & \textbf{40} & \textbf{41} & \textbf{42} & \textbf{43} & \textbf{44} & \textbf{45} & \textbf{46} & \textbf{47} \\
\midrule
\multirow{3}{*}{\textit{$General$}} & pre-patch & 0.017 &   &   &   &   &   &   &   &   &   &   &   &   &   &   &   &   &   &   &   &   &   &   &   &   &   &   \\
 & patch$_{t1}$ & 0.018 & 0.020 & 0.019 & 0.022 & 0.022 & 0.024 & 0.024 & 0.026 & 0.030 & 0.029 & 0.034 & 0.030 & 0.028 & 0.032 & 0.038 & 0.036 & 0.051 & 0.053 & 0.039 & 0.048 & 0.053 & 0.057 & \underline{0.070} & \textbf{0.073} & 0.059 & \underline{0.064} & 0.052 \\
 & patch$_{t2}$ & 0.012 & 0.008 & 0.011 & 0.007 & 0.011 & 0.006 & 0.009 & 0.007 & 0.008 & 0.009 & -0.000 & 0.002 & 0.004 & -0.000 & -0.007 & -0.004 & -0.007 & -0.013 & -0.004 & -0.016 & -0.014 & -0.024 & -0.026 & \textbf{-0.038} & \underline{-0.031} & \underline{-0.028} & -0.020 \\
\midrule
\multirow{3}{*}{\textit{$Absolute$}} & pre-patch & 0.270 &    &    &    &    &    &    &    &    &    &    &    &    &    &    &    &    &    &    &    &    &    &    &    &    &    &    \\
 & patch$_{t1}$ & 0.270 & 0.272 & 0.270 & 0.272 & 0.272 & 0.272 & 0.270 & 0.273 & 0.274 & 0.274 & 0.273 & 0.273 & 0.274 & 0.276 & 0.275 & 0.272 & 0.280 & 0.279 & 0.274 & 0.281 & 0.281 & 0.276 & \textbf{0.290} & \underline{0.281} & 0.275 & 0.278 & \underline{0.286} \\
 & patch$_{t2}$ & 0.269 & 0.272 & 0.270 & 0.272 & 0.271 & 0.273 & 0.272 & 0.273 & 0.272 & 0.272 & 0.272 & 0.273 & 0.276 & 0.275 & 0.272 & 0.270 & 0.278 & 0.275 & 0.271 & \underline{0.279} & 0.278 & 0.276 & \underline{0.281} & 0.277 & 0.275 & 0.278 & \textbf{0.287} \\
\midrule
\multirow{3}{*}{\textit{$General_{p3}$}} & pre-patch & 0.009 &   &   &   &   &   &   &   &   &   &   &   &   &   &   &   &   &   &   &   &   &   &   &   &   &   &   \\
 & patch$_{t1}$ & 0.008 & 0.010 & 0.013 & 0.016 & 0.016 & 0.019 & 0.017 & 0.021 & 0.023 & 0.020 & 0.022 & 0.021 & 0.021 & 0.025 & 0.029 & 0.032 & 0.042 & 0.049 & 0.034 & 0.041 & 0.049 & 0.055 & \underline{0.069} & \textbf{0.073} & 0.063 & \underline{0.064} & 0.047 \\
 & patch$_{t2}$ & 0.004 & -0.003 & 0.002 & 0.001 & 0.006 & -0.003 & 0.003 & -0.002 & -0.001 & 0.002 & -0.010 & -0.006 & -0.000 & -0.012 & -0.014 & -0.011 & -0.015 & -0.024 & -0.014 & -0.028 & -0.023 & -0.036 & -0.036 & \textbf{-0.050} & \underline{-0.038} & \underline{-0.038} & -0.026 \\
\midrule
\multirow{3}{*}{\textit{$Absolute_{p3}$}} & pre-patch & 0.357 &   &   &   &   &   &   &   &   &   &   &   &   &   &   &   &   &   &   &   &   &   &   &   &   &   &   \\
 & patch$_{t1}$ & 0.359 & 0.360 & 0.358 & 0.361 & 0.360 & 0.362 & 0.359 & 0.358 & \underline{0.366} & 0.360 & 0.359 & 0.361 & 0.366 & 0.366 & 0.362 & \underline{0.367} & 0.362 & 0.360 & 0.365 & 0.366 & 0.365 & 0.354 & \textbf{0.371} & 0.362 & 0.355 & 0.348 & 0.364 \\
 & patch$_{t2}$ & 0.359 & 0.358 & 0.357 & 0.363 & 0.358 & 0.359 & 0.360 & 0.355 & \underline{0.366} & 0.360 & 0.356 & 0.358 & 0.362 & 0.360 & 0.359 & \textbf{0.369} & 0.363 & 0.360 & 0.362 & 0.366 & 0.365 & 0.350 & \underline{0.368} & 0.366 & 0.358 & 0.346 & 0.365 \\
\midrule
\multirow{3}{*}{\textit{$General_{p4}$}} & pre-patch & 0.020 &   &   &   &   &   &   &   &   &   &   &   &   &   &   &   &   &   &   &   &   &   &   &   &   &   &   \\
 & patch$_{t1}$ & 0.021 & 0.025 & 0.024 & 0.026 & 0.027 & 0.026 & 0.023 & 0.030 & 0.035 & 0.030 & 0.038 & 0.032 & 0.031 & 0.036 & 0.044 & 0.041 & 0.054 & 0.061 & 0.046 & 0.056 & 0.058 & 0.067 & \underline{0.081} & \textbf{0.084} & 0.066 & \underline{0.073} & 0.056 \\
 & patch$_{t2}$ & 0.019 & 0.012 & 0.016 & 0.012 & 0.018 & 0.009 & 0.010 & 0.009 & 0.011 & 0.012 & 0.004 & 0.004 & 0.010 & 0.004 & 0.001 & -0.000 & -0.005 & -0.012 & -0.005 & -0.016 & -0.015 & -0.019 & -0.023 & \textbf{-0.040} & \underline{-0.030} & \underline{-0.026} & -0.013 \\
\midrule
\multirow{3}{*}{\textit{$Absolute_{p4}$}} & pre-patch & 0.358 &   &   &   &   &   &   &   &   &   &   &   &   &   &   &   &   &   &   &   &   &   &   &   &   &   &   \\
 & patch$_{t1}$ & 0.358 & 0.362 & 0.359 & 0.361 & 0.359 & 0.361 & 0.359 & 0.361 & 0.368 & 0.364 & 0.362 & 0.361 & 0.364 & 0.368 & 0.362 & \underline{0.371} & 0.365 & 0.361 & \underline{0.368} & 0.366 & 0.368 & 0.359 & \textbf{0.376} & 0.363 & 0.361 & 0.352 & 0.364 \\
 & patch$_{t2}$ & 0.360 & 0.363 & 0.357 & 0.362 & 0.356 & 0.360 & 0.357 & 0.357 & 0.366 & 0.362 & 0.357 & 0.360 & 0.364 & 0.363 & 0.359 & \underline{0.371} & 0.363 & 0.358 & 0.365 & \underline{0.367} & 0.364 & 0.355 & \textbf{0.372} & 0.364 & 0.362 & 0.346 & 0.364 \\
\bottomrule
\end{tabular}%
}
\caption{Probability gaps across layers before patching (pre-patch), after patching $t_1$ (patch$_{t1}$), and after patching $t_2$ (patch$_{t2}$). For general patching, we measure the different probability difference of $t_1$ to $t_2$ (e.g. $diff = P(t_1) - P(t_2)$). The results of patching with $t_2$ in these cases are shown as a sanity check (the gap is expected to decrease or inverse as it would increase $P(t_2)$ and decrease $P(t_1)$). For this reason, the best results for these cases are the lowest values.}
\label{tab:probability_gaps}
\end{table*}

\end{document}